\documentclass[12pt]{article}

\usepackage[top=1in, bottom=1in, left=1in, right=1in]{geometry}
\usepackage[rflt]{floatflt}
\usepackage{graphicx}
\usepackage{times}

\usepackage[authoryear]{natbib}

\usepackage{enumitem}

\usepackage{xcolor} 

\usepackage{url}

\usepackage{graphicx}
\usepackage{subfigure}
\usepackage{algorithm}
\usepackage{algorithmic}
\usepackage{booktabs}

\usepackage{url}
\usepackage{verbatim}
\usepackage{amsmath}
\usepackage{amsfonts}
\usepackage{amssymb}

\DeclareMathOperator*{\argmax}{arg\,\!max}

\newtheorem{definition}{Definition}

\newcommand{\fix}[1]{}

\usepackage{amssymb}
\usepackage{amsfonts}
\usepackage{amsmath}
\providecommand{\joint}[1]{\mathbf{#1}}

\newcommand{\set}[1]{\mathbb{#1}}

\newcommand{\Exp}{\mathbb{E}}

\providecommand{\argsA}[2]{ {#1}_{#2} }             
\providecommand{\argsI}[2]{ {#1}^{#2} }           
\providecommand{\argsT}[2]{ {#1}_{#2} }             

\providecommand{\argsAT}[3]{ {#1}_{#2,#3} }

\providecommand{\agentS}{\set{I}}

\providecommand{\stateS}{\set{S}}
\providecommand{\sI}[1]{\argsI {s}   {#1}}   

\providecommand{\discount}{\gamma}

\providecommand{\hor}{\mathcal{H}}

\providecommand{\nrA}{n} 

\providecommand{\ts}{t}        

\providecommand{\bSymbol}           {{b}}
\providecommand{\bO}               { {\argsT\bSymbol{0}}}               

\providecommand{\POL}{\pi}

\providecommand{\jpol}      {\joint{\POL}}	 

\providecommand{\polA}[1]   {\argsA{\POL}{#1}}	
\providecommand{\poli}   {\polA i}  

\providecommand{\AOH}{h} 
\providecommand{\AOHS}{\set{H}} 

\providecommand{\JAOH}{\joint{\AOH}{}} 
\providecommand{\JAOHS}{\joint{\AOHS}{}} 
\providecommand{\jh}        {\JAOH}
\providecommand{\jhT}[1]{\argsT {\jh}   {#1}}   
\providecommand{\jhS}       {\JAOHS}

\providecommand{\aoHistA}[1]    {\argsA   {\AOH}{#1}} 
\providecommand{\hi    {\aoHistA i}}  
\providecommand{\hiT}[1]{\argsAT {\AOH} {i} {#1}} 
\providecommand{\hAOHS}[1]{\argsA {\AOHS}  {#1}} 
\providecommand{\aoHistAT}[2]   {\argsAT  {\AOH}{#1}{\,#2}}

\providecommand{\AC}{a}                 
\providecommand{\ACS}{\set{A}}                     

\providecommand{\ja}    {\joint {\AC}}          
\providecommand{\jaS}   {\joint {\ACS}}         
\providecommand{\jaT}[1]{\argsT {\ja}   {#1}}   

\providecommand{\aA}[1] {\argsA {\AC}   {#1}} 
\providecommand{\ai    {\aA i}} 
\providecommand{\aiT}[1]{\argsAT {\AC} {i} {#1}} 
\providecommand{\aAS}[1] {\argsA {\ACS}  {#1}} 
\providecommand{\aAT}[2]{\argsAT{\AC}   {#1}{#2}}

\providecommand{\OB}{o}                
\providecommand{\OBS}{\set{O}}

\providecommand{\jo}    {\joint {\OB}}      
\providecommand{\joS}   {\joint {\OBS}}    
\providecommand{\joT}[1]{\argsT {\jo}   {#1}} 

\providecommand{\oA}[1] {\argsA {\OB}   {#1}} 
\providecommand{\oi    {\oA i}} 
\providecommand{\oiT}[1]{\argsAT {\OB} {i} {#1}} 
\providecommand{\oAS}[1]{\argsA {\OBS}  {#1}}   
\providecommand{\oASi    {\oA i}} 

\providecommand{\jQ}    {\joint {Q}}
\providecommand{\jQpol}    {\joint {Q^{\jpol}}}
\providecommand{\QA}[1] {\argsA {Q}   {#1}} 
\providecommand{\Qi}    {\QA i} 
\providecommand{\jQ}    {\joint {Q}}
\providecommand{\jQmodel}    {\hat{\joint {Q}}}

\providecommand{\jV}    {\joint {V}}  

\providecommand{\VA}[1] {\argsA {V}   {#1}} 
\providecommand{\Vi}    {\VA i} 
\providecommand{\jVmodel}    {\hat{\joint {V}}}
\providecommand{\Vimodel}    {\hat \Vi} 

\providecommand{\jA}    {\joint {A}}  
\providecommand{\Ai}    {\argsA A i} 
\providecommand{\jAmodel}    {\hat{\joint {A}}}

\providecommand{\rT}[1]{\argsT {r}   {#1}} 

\providecommand{\nablasub}[1]{\nabla_{\! {#1}}}
\newcommand\nablai{\nabla_{\! \ppi}}
\providecommand{\thetai} {\argsA {\theta}  {i}} 

\providecommand{\valparam}{\theta}
\providecommand{\polparam}{\psi}

\providecommand{\ppi}{\argsA {\polparam} {i}}

\title{\vspace{-20pt}An Introduction to \\Centralized Training for Decentralized Execution in \\Cooperative Multi-Agent Reinforcement Learning}
\author{Christopher Amato, Northeastern University}

\begin{document}

\maketitle

\tableofcontents

\pagebreak

Multi-agent reinforcement learning (MARL) has exploded in popularity in recent years. Many approaches have been developed but they can be divided into three main types: centralized training and execution (CTE), centralized training for decentralized execution (CTDE), and Decentralized training and execution (DTE). 

CTE methods assume centralization during training and execution (e.g., with fast, free and perfect communication) and have the most information during execution. That is, the actions of each agent can depend on the information from all agents. As a result, a simple form of CTE can be achieved by using a single-agent RL method with centralized action and observation spaces (maintaining a centralized action-observation history for the partially observable case). CTE methods can potentially outperform the decentralized execution methods (since they allow centralized control) but are less scalable as the (centralized) action and observation spaces scale exponentially with the number of agents. CTE is typically only used in the cooperative MARL case since centralized control implies coordination on what actions will be selected by each agent. 
CTDE methods are the most common as they can use centralized information during training but execute in a decentralized manner---using only information available to that agent during execution. CTDE is the only paradigm that requires a separate training phase where any available information (e.g., other agent policies, underlying states) can be used. 
As a result, they can be more scalable than CTE methods, do not require communication during execution, and can often perform well.  CTDE fits most naturally with the cooperative case, but can be potentially applied in competitive or mixed settings depending on what information is assumed to be observed. 
Decentralized training and execution methods make the fewest assumptions and are often simple to implement. In fact, any single-agent RL method can be used for DTE by just letting each agent learn separately. Of course, there are pros and cons to such approaches~\citep{DTE}. It is worth noting that DTE is required if no centralized training phase is available (e.g., though a centralized simulator), requiring all agents to learn during online interactions without prior coordination.   
DTE methods can be applied in cooperative, competitive, or mixed cases. 

MARL methods can be further broken up into value-based and policy gradient methods. Value-based methods (e.g., Q-learning) learn a value function and then choose actions based on those values. Policy gradient methods learn an explicit policy representation and attempt to improve the policy in the direction of the gradient. Both classes of methods are widely used in MARL. 

This text is an introduction to CTDE MARL. It is meant to explain the setting, basic concepts, and common methods. It does not cover all work in CTDE MARL as the subarea is quite extensive. I have included work that I believe is important for understanding the main concepts in the subarea and apologize to those that I have omitted. 

I will first give a brief description of the cooperative MARL problem in the form of the Dec-POMDP.  Then, I present an overview of CTDE and  the two main classes of CTDE methods: value function factorization methods and centralized critic actor-critic methods. Value function factorization methods include the well-known VDN \citep{VDN}, QMIX \citep{QMIX}, and QPLEX \citep{QPLEX} approaches, while centralized critic methods include MADDPG \citep{MADDPG}, COMA \citep{COMA}, and MAPPO \citep{MAPPO}.  Finally, I discuss other forms of CTDE such as adding centralized information to decentralized (i.e., independent) learners (such as parameter sharing) and decentralizing centralized solutions.

The basics of reinforcement learning (in the single-agent setting) are not presented in this text. Anyone interested in RL should read the book by \cite{SuttonBarto18}. Similarly, for a broader overview of MARL, the recent book by Albrecht, Christianos and Sch\"afer is recommended \citep{marl-book}.

\section{The cooperative MARL problem: The Dec-POMDP}
\label{sec:decpomdp}

\begin{figure}[t]
\centering
\includegraphics[height=0.2\linewidth]{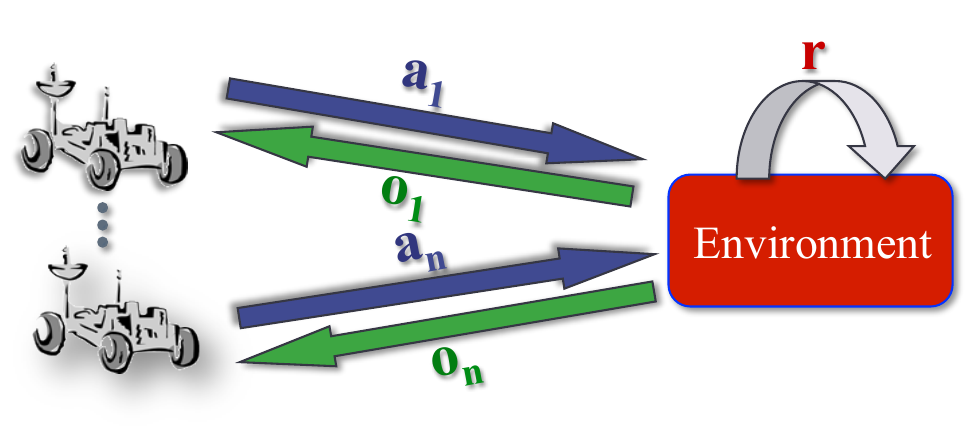}\label{fig:decpomdp}
\caption{A depiction of cooperative MARL---a Dec-POMDP.}
\label{fig:SARPicture}
\end{figure}

The cooperative multi-agent reinforcement learning (MARL) problem can be represented as a Dec-POMDP \citep{Book16,Bernstein02MOR}. 
Dec-POMDPs generalize POMDPs \citep{Kaelbling98AI} (and MDPs \citep{Puterman94}) to the multi-agent, decentralized setting. 
As depicted in Figure~\ref{fig:decpomdp}, multiple agents  operate under uncertainty based on partial views of the world, with execution unfolding over time. At each step, every agent chooses an action (in parallel) based purely on locally observable information, resulting in each agent obtaining  an observation and the team obtaining a joint reward. The shared reward
function makes the problem cooperative, but their local views mean that execution is decentralized. 

Formally, a  Dec-POMDP is defined by tuple $ \langle  \agentS, \stateS, \{\aAS i\}, T, R, \{\oAS i\}, O, \hor, \discount \rangle $. For simplicity, I define the finite version but it easily extends to the continuous case: 
\begin{itemize}
\item 
$\agentS$ is a finite set of agents of size $|\agentS|=\nrA$;
\item 
$\stateS$ is a finite set of states with designated initial state
distribution $\bO$;\footnote{Some papers refer to other information such as joint observations or histories as `state.' For clarity, I will only use this term to refer to the true underlying state.}
\item  
$\aAS i$ is a finite set of actions for each agent $i$ with $\jaS=\times_i \aAS i$ the set of joint actions; 
\item  
$T$ is a state transition probability function, $T: \stateS \times \jaS \times \stateS \to [0,1]$, that specifies the
probability of transitioning from state $s \in \stateS$ to $s' \in \stateS$ when the actions $\ja \in \jaS$ are taken by the agents (i.e., $T(s,\ja,s') = \Pr(s' | \ja,s)$);
\item 
 $R$ is a reward function: $R: \stateS \times \jaS \to \mathbb{R}$, the immediate reward for being in state $s \in \stateS$ and taking the
actions $\ja \in \jaS$;
\item
 $\oAS i$ is a finite set of observations for each agent, $i$, with $\joS=\times_i \oAS i$ the set of joint observations;
\item
  $O$ is an observation probability function: $O: \joS \times \jaS \times \stateS \to [0,1]$, the
probability of  seeing  observations $\jo \in \joS$  given 
actions $\ja \in \jaS$  were taken and resulting in state $s' \in \stateS$ (i.e.,  $O(\ja,s',\jo) = \Pr(\jo | \ja,s')$);
\item 
$\hor$ is the number of steps until termination, called the horizon;
\item and $\discount \in [0,1]$ is the discount factor. 
\end{itemize}
\vspace{-5pt}
I also include both the horizon and discount in the definition but, typically, only one is used. When $\hor$ is finite, $\gamma$
 can be set to 1 and when $\hor=\infty$,  $\discount \in [0,1)$.\footnote{The episodic case \citep{SuttonBarto18} is typically indefinite-horizon with termination to a set of terminal states with probability 1 and the infinite-horizon case is sometimes called `continuing.'}

A solution to a Dec-POMDP is a \emph{joint policy}, denoted $\jpol$---a set of policies, one for each agent, each of which is denoted $\poli$. Because the state is not directly observed, it is typically beneficial for each agent to remember a history of its observations (and potentially actions). Then,
a \emph{local policy}, $\poli$, for an agent is a mapping from local action-observation histories to actions or probability distributions over actions. 
A \emph{local deterministic policy} for agent $i$, $\poli(\hi)=\ai$, maps $\hAOHS i \to \aAS i$, where $\hAOHS i$ is the set of \emph{local observation histories}, $\hi = \{\aiT 0, \oiT 0, \ldots, \aiT {\ts-1}, \oiT {\ts-1} \}$\footnote{Sometimes, $\oiT 0$ is also included as an observation generated by the initial state distribution}, by agent $i$ up to the current time step, $t$. 
Note that histories have implicit time steps due to their length which we do not include in the notation (i.e., we always assume a history starts on the first time step and the last time step is defined by the number of action-observation pairs).  
We can denote the \emph{joint histories} for all agents at a given time step as $\jh=\langle \aoHistA 1, \ldots, \aoHistA \nrA \rangle$. 
A \emph{joint deterministic policy} is denoted $\pi(\jh)=\ja=\langle\polA 1(\aoHistA 1), \dots, \polA \nrA(\aoHistA \nrA)  \rangle=\langle\aA 1, \ldots,  \aA \nrA \rangle$. 
A \emph{stochastic local policy} for agent $i$ is $\poli(\ai| \hi)$, representing the probability of choosing action $\ai$ in history $\hi$.  
A \emph{joint stochastic policy} is denoted  $\jpol(\ja| \jh)=\prod_{i \in \agentS} \poli(\ai| \hi)$. 
Because one policy is generated for each agent and these policies depend only on local observations, they operate in a decentralized manner. 

Many researchers just use observation histories (without including actions), which is sufficient for deterministic policies but may not be for stochastic policies \citep{Book16}. 
Deterministic policies will be used in the value-based methods in Section \ref{sec:ctde:vff} while stochastic policies will be used in the policy gradient gradient methods in Section \ref{sec:ctde:pg}.
There always exists an optimal deterministic joint policy in Dec-POMDPs \citep{Book16}, but stochastic (or continuous) policies are needed for policy gradient methods.

The value of a joint policy, $\pi$, at joint history $\jh$ can be defined for the case of discrete states and observations as
\begin{equation}
 \jV^\jpol(\jh)= \sum_{s} P(s|\jh,\bO) \Big[R(s,\ja)+\discount \sum_{s'}P(s'|\ja,s)\sum_{\jo}P(\jo|\ja,s') \jV^\jpol(\jh\ja\jo)\Big]\Big|_{\ja=\jpol(\jh)}
\label{eq:BellmanV}
\end{equation}
where $P(s|\jh,\bO)$ is the probability of state $s$ after observing joint history $\jh$ starting from state distribution $\bO$ and $\ja=\jpol(\jh)$ is the joint action taken at the joint history. Also, $\jh\ja\jo$ represents $\jh'$, the joint history after taking joint action $\ja$ in joint history $\jh$ and observing joint observation $\jo$. 
In the RL context, algorithms do not iterate over states and observations to explicitly calculate this expectation but approximate it through sampling. 
For the finite-horizon case, $\jV^\jpol(\jh)=0$ when the length of $\jh$ equals the horizon $\hor$, showing the value function includes the time step from the history.

We will often want to evaluate policies starting from the beginning---starting at the initial state distribution before any action or observation. Starting from this initial, null history we denote the value of a policy as $\jV^\jpol(\jh_0)$. 
An  \emph{optimal joint policy} beginning at  $\jh_0$ is 
$\argmax_{\jpol} \jV^{\jpol}(\jh_0)$, where the $\argmax$ here denotes enumeration over decentralized policies. 
The optimal joint policy is then the set of local policies for each agent that provides the highest value, which is denoted $\jV^*$.

Reinforcement learning methods often use history-action values, $\jQ(\jh,\ja)$, rather than just history values $\jV(\jh)$. $\jQ^{\jpol}(\jh,\ja)$ is the value of choosing joint action $\ja$ at joint history $\jh$ and then continuing with policy $\jpol$, 
\begin{equation}
 \jQ^\jpol(\jh, \ja)= \sum_{s} P(s|\jh,\bO) \Big[R(\ja,s)+\discount \sum_{s'}P(s'|\ja,s)\sum_{\jo}P(\jo|\ja,s') \jQ^\jpol\big(\jh\ja\jo, \ja')|_{\ja'=\jpol(\jh\ja\jo)}\big)\Big],
 \label{eq:BellmanQ}
\end{equation}
while $\jQ^*(\jh,\ja)$ is the value of choosing action $\ja$ at history $\jh$ and then continuing with the optimal policy, $\jpol^*$.

\paragraph{Policies that depend only on a single observation}
\label{sec:singleobs}
It is somewhat popular to define policies that depend only on a single observation rather than the whole observation history. 
That is, $\poli: \oAS i \to \aAS i$, rather than $\hAOHS i \to \aAS i$. This type of policy is often referred to as \emph{reactive} or \emph{Markov} since it just depends on (or reacts from) the past observation. These reactive policies are typically not desirable since they can be arbitrarily worse than policies that consider history information \citep{Murphy00} but can perform well or even be optimal in simpler subclasses of Dec-POMDPs \citep{Goldman04JAIR}. In general, reactive policies reduce the complexity of finding a solution (since many fewer policies need to be considered) but only perform well in limited  cases such as when the problem is not really partially observable or when the observation history isn't helpful for decision-making.

\paragraph{The fully observable case}
\label{sec:fullobs}
It is worth noting that if the cooperative problem is fully observable, it becomes much simpler. 
Specifically, a multi-agent MDP (MMDP) can be defined by the tuple $ \langle  \agentS, \stateS, \{\aAS i\}, T, R, \hor, \discount \rangle $\citep{Boutilier96TARK}. Each agent can observe the full state in this case. In the CTDE context, solving an MMDP can be done by standard MDP methods where the action space is the joint action space, $\jaS=\times_i \aAS i$. The resulting policy, $\stateS \to \jaS$, can be trivially decentralized as $\stateS \to \aAS i$ for each agent $i$. A number of methods have been developed for learning in MMDPs (and other multi-agent models) \citep{Busoniu08IEEE_SMC}. 

Finding solutions in Dec-POMDPs is much more challenging. 
This can even be seen by the (finite-horizon) complexity for finite problems---optimally solving an MDP is P-complete (polynomial in the size of the system) \citep{papadimitriou1987complexity,Littman95} and 
optimally solving a POMDP is PSPACE (essentially exponential in the actions and observations), 
\citep{papadimitriou1987complexity}, while optimally solving a Dec-POMDP is NEXP-complete (essentially doubly exponential) \citep{Bernstein02MOR}. 
This can be intuitively understood by considering the simplest possible solution method for each class of problems. 
For an MDP, you could search all policies mapping states to (joint) actions: $|\jaS|^{|\stateS|}$ of them. 
For a POMDP, you need to consider histories so centralized control (mapping joint histories to joint actions) would result in searching over $|\jaS|^{|\joS|^\hor}$ possibilities.\footnote{Since we are assuming deterministic policies, the actions are fixed but we would still have to search over all possible observation histories up to the given horizon, resulting in $\joS^{\hor}$ possible histories. }
In the Dec-POMDP case, each agent would have $|\aAS i|^{|\oAS i|^\hor}$ possible policies, so the possible number of possible joint policies would be $|\aAS i|^{|\oAS i|^{\hor^{|\agentS|}}}$ if all agents have the same number of actions and observations as agent $i$. 
Of course, state-of-the-art methods are more sophisticated than this and the complexity is for the worst case. Most real-world problems are intractable to solve optimally anyway.  
Regardless, these numbers make it clear that searching the joint space of history-based policies is enormous and the resulting partially observable problem is much more challenging than the fully observable case. 

\section{CTDE overview}

The idea of centralized training for decentralized execution is quite vague. The general idea of CTDE the Dec-POMDP case originated (with the Dec-POMDP itself) in the planning literature \citep{Bernstein02MOR}, where \emph{planning} would be centralized but execution was decentralized. This idea goes back further to \emph{team} decision making more generally since it is natural to think of deriving a solution for the team as a whole and then assigning corresponding parts to the team members \citep{Marschak55,Radner62,Ho80}.
This concept was carried over to the reinforcement learning case and has come to mean that there is an `offline'\footnote{Offline does not refer to having a fixed training set \citep{OfflineRL} but rather in a different setting than what is used for execution (which would be considered online).} training phase where some amount of centralization is allowed (such as in \citep{Kraemer16}). After this centralized training phase is complete, agents must then act in a decentralized manner based on their own histories (as given by the Dec-POMDP definition above). Any information that would not be available during decentralized execution can be used in the centralized training phase ranging from neural network parameters, policies of the other agents, a joint value estimate, etc. Nevertheless, CTDE has never been formally defined (to the best of my knowledge) and I will not do so here. I will take a wide view of CTDE and consider it to include any centralized information or coordination during learning---anything that is not available during decentralized execution.  

CTDE methods for Dec-POMDPs vary widely. 
Some methods add centralized information to DTE methods. These approaches should be scalable but use limited centralized information. 
Other methods are at the other extreme, where they learn a centralized policy and then attempt to decentralize it so it can be executed without the centralized information. 
Most CTDE methods fall somewhere in between these two ideas. The main class of value-based CTDE methods learns a set of decentralized Q-functions (one for each agent) by factoring a joint value function, as in value function factorization methods.  
The main class of policy gradient CTDE methods learns a centralized critic that approximates the joint value function for all agents and that centralized critic is used to update a set of decentralized actors (one per agent). 
I discuss these ideas in more detail below.

\section{Value function factorization methods}
\label{sec:ctde:vff}

There is a long history of learning factored value functions in multi-agent reinforcement learning, but many previous methods were focused on more efficiently learning a joint value function (e.g., \citep{Kok06JMLR}). Other methods have also been developed that share Q-functions during training to improve coordination \citep{Schneider99}. 
Modern value function factorization methods learn Q-functions for each agent by combining them into a joint Q-function and calculating a loss based on the joint Q-function. In this way, the joint Q-function is factored into local, decentralized Q-values that can be used during decentralized execution. 
This idea is appealing because a joint Q-function can be learned to approximate the one in Equation \ref{eq:BellmanQ} but agents can choose actions based on their decentralized Q-values. Additional details and the most popular methods are below.

\subsection{Background on value-based RL}
\label{sec:value:back}

I will first provide background on the value-based methods that the value function factorization methods build upon.  Therefore, I will first introduce Deep Q-Networks (DQN), and the extension to partial observability, DRQN.

\begin{figure}
\hfill
\begin{centering}
\subfigure[DQN]{
\includegraphics[scale=0.6]{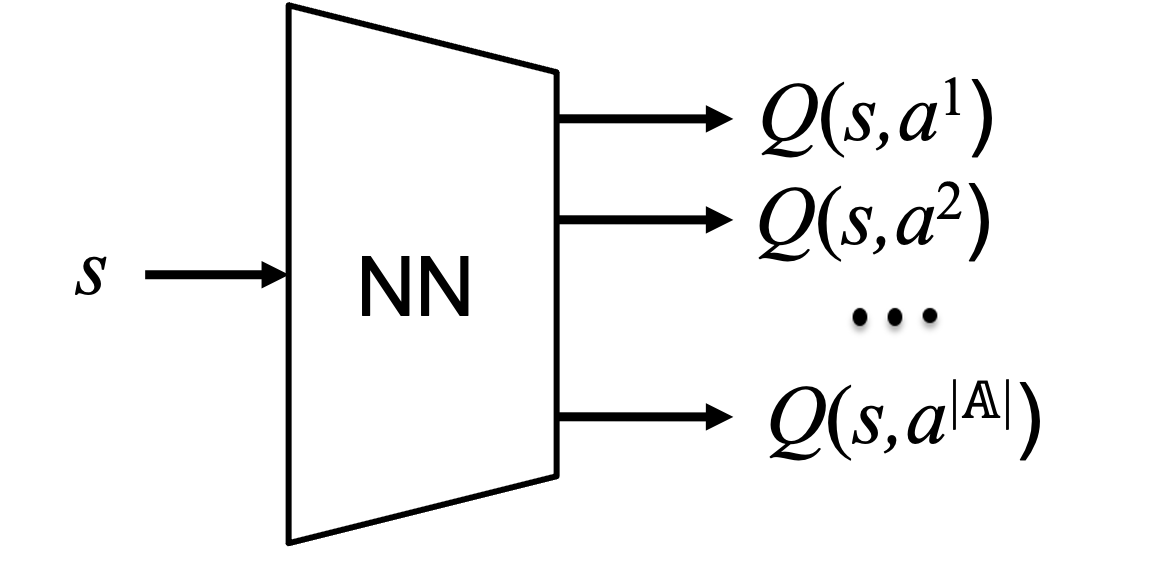} \label{fig:DQN}
}
\subfigure[DRQN]{
\includegraphics[scale=0.5]{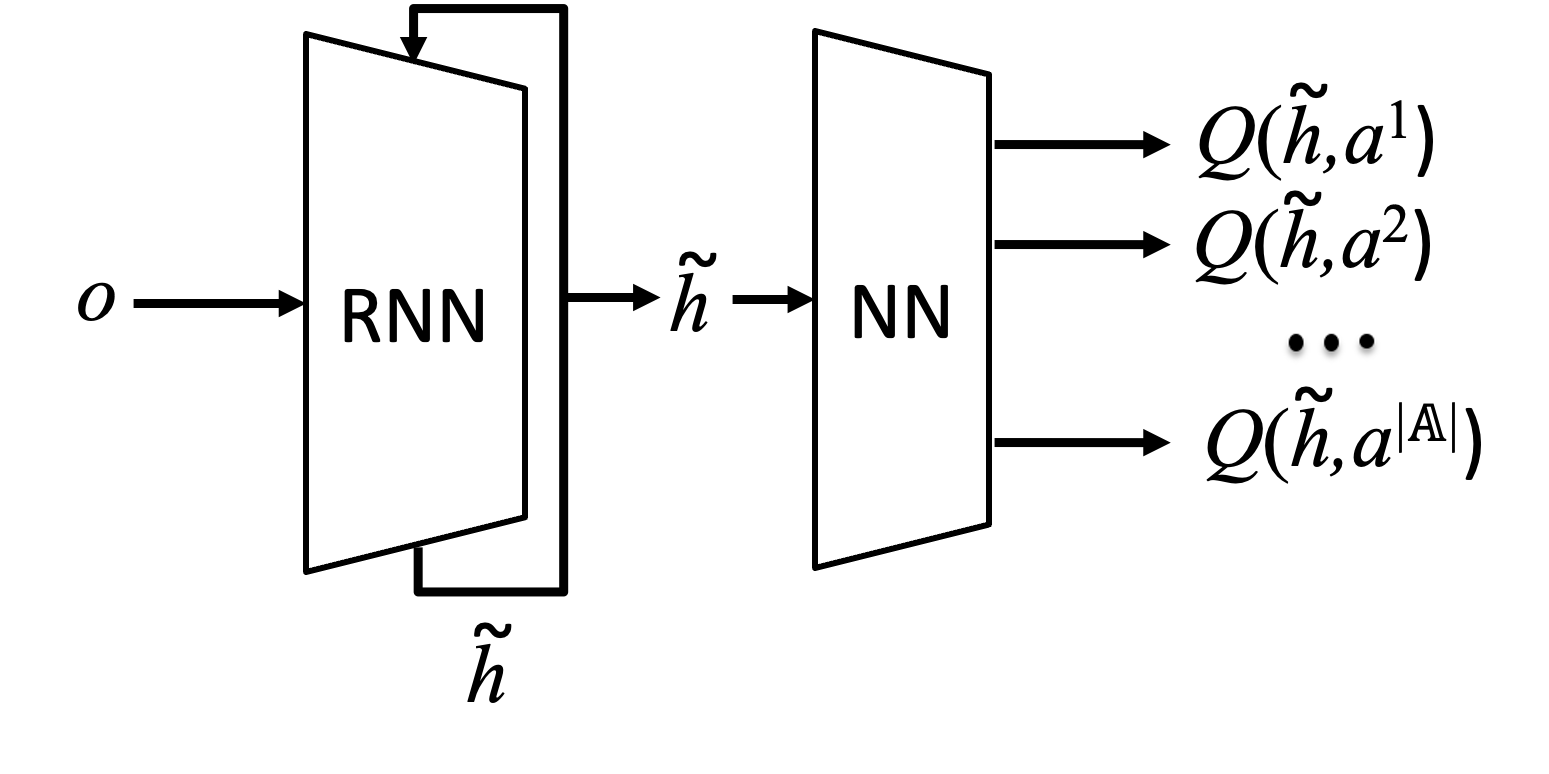}
\label{fig:DRQN}
}
\hfill
\end{centering}
\caption{DQN and DRQN diagrams.}
\end{figure}

\paragraph{Deep Q-networks (DQN)}\citep{DQN} is an extension of Q-learning~\citep{Watkins1992} to include a neural net as a function approximator. 
Since DQN was designed for the fully observable (MDP) case, it learns $Q_\theta(s,a)$, parameterized with $\theta$ (i.e., $\theta$ represents the parameters of the neural network), by minimizing the loss:\\[-7pt]
\begin{equation}
    \mathcal{L}(\theta)=\mathbb{E}_{<s, a, r, s'>\sim\mathcal{D}}\Big[\big(y - Q_{\theta}(s, a)\big)^2 \Big] \text{, where\,\, } y=r + \gamma\max_{a'}Q_{\theta^-}(s',a')
    \label{eq:dqn}
\end{equation}
which is just the squared TD error---the difference between the current estimated value, $Q_{\theta}(s, a)$, and the new value gotten from adding the newly seen reward to the previous Q-estimate at the next state, $Q_{\theta^-}(s',a')$. Because learning neural networks can be unstable, a separate target action-value function $Q_{\theta^-}$ and an experience replay buffer $\mathcal{D}$ \citep{Lin92} are implemented to stabilize learning. The target network is an older version of the Q-estimator that is updated periodically with $\langle s,a,r,s'\rangle$ sequences stored in the experience replay buffer and single $\langle s,a,r,s'\rangle$ tuples are i.i.d.~sampled for updates. 
As shown in Figure \ref{fig:DQN}, the neural network (NN) outputs values for all actions ($\AC \in \ACS$) to make maximizing over all states possible with a single forward pass (rather than iterating through the actions). 

\paragraph{Deep recurrent Q-networks (DRQN)}\citep{DRQN} extends DQN to handle partial observability, where some number of recurrent layers (e.g., LSTM \citep{LSTM}) are included to maintain an internal hidden state which is an abstraction of the history (as shown as $\tilde h$ in Figure \ref{fig:DRQN}). Because the problem is partially observable, $o,a,r,o'$ sequences are stored in the replay buffer during execution. The update equation is very similar to that of DQN:
\begin{equation}
    \mathcal{L}(\theta)=\mathbb{E}_{<h,a,r,o>\sim\mathcal{D}}\Big[\big(y - Q_{\theta}(h, a)\big)^2 \Big] \text{, where\,\, } y=r + \gamma\max_{a'}Q_{\theta^-}(h',a')
    \label{eq:drqn}
\end{equation}
but since the recurrent neural network (RNN) is used, the internal state of the RNN can be thought of as a history representation. 
RNNs are trained sequentially, updating the internal state at each time step. As a result, the figure shows only $o$ as the input but this assumes the internal state $\tilde h$ has already been updated using the history up to this point $h^{t-1}$. 
Therefore, rather than just sampling single updates ($o,a,r,o'$) i.i.d, like in DQN, histories are sampled from the buffer. The internal state can be updated incrementally and the Q-values learned starting from the first time step and going until the end of the history (i.e., the horizon or episode). 
An example of training using DRQN is shown in Algorithm \ref{alg:VDN}.
Technically, a whole history (e.g., episode) should be sampled from the replay buffer to train the recurrent network but it is common to use a fixed history length (e.g., sample $o,a,o',r$ sequences of length 10). 
Also, as mentioned above, many implementations just use observation histories rather than full action-observation histories. 
Note that I will still write $h$ rather than $\tilde h$ in the equations (e.g., $Q_{\theta}(h, a)$ above) to be more general. 
The resulting equations are not restricted to recurrent models (e.g., use a non-recurrent representation such as the full history or a Transformer-based representation \citep{DTQN,Ni23}). 
With the popularity of DRQN, it has become common to add recurrent layers to standard (fully observable) deep reinforcement learning methods when solving partially observable problems.

\subsection{VDN}

\begin{figure}
\begin{centering}

\includegraphics[scale=0.7]{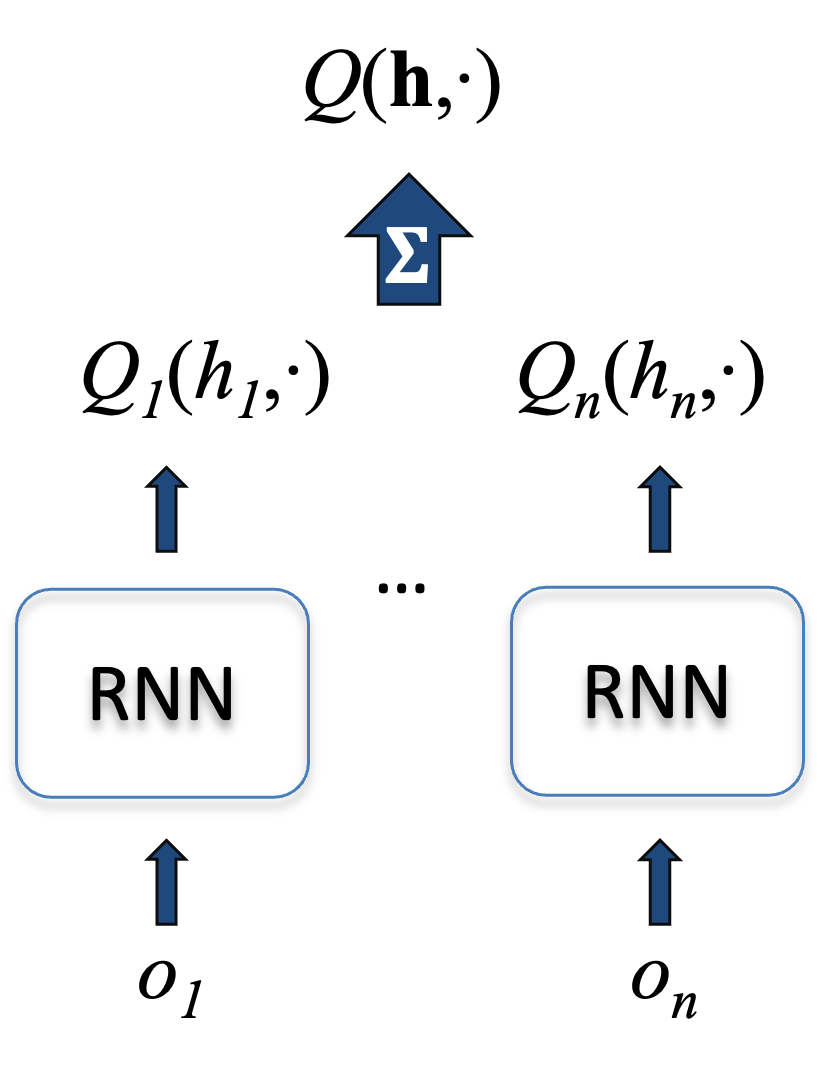}

\end{centering}
\caption{VDN architecture diagram}

\label{fig:VDN}
\end{figure}

Value decomposition networks (VDN) \citep{VDN} began the popular trend in value decomposition methods for solving multi-agent reinforcement learning. 
The main contribution in VDN is a decomposition of the joint Q-function into additive Q-functions per agent.\footnote{The paper also uses dueling \citep{DuelingDQN} and investigates forms of weight sharing and communication but we view these as orthogonal to the main value decomposition contribution. } This idea is relatively simple but very powerful. It allows training to take place in a centralized setting but agents can still execute in a decentralized manner because they learn individual Q-functions per agent, which can then be maxed over to select an action. 

In particular, it assumes the following approximate factorization of the Q-function:
\begin{equation}
\jQ(\jh, \ja) \approx \sum_{i \in \agentS}^\nrA \Qi(\hi,\ai)
\end{equation}
That is, the joint Q-function is approximated by a sum over each agent's individual Q-function. This is a form of factorization---the joint Q-function is factored as a sum of the local $\Qi$ functions \citep{Guestrin01}. 
It is worth noting that each $\Qi$ is technically a utility and not a value function since it is not estimating a particular expected return but can be any arbitrary value in the sum. 

The architecture diagram is given in Figure \ref{fig:VDN}. Each agent learns an individual set of Q-values that depend only on local information. These Q-networks take the current observation as input to a recurrent network, which updates the history representation and then outputs Q-values for that agent. During training, these individual Q-values are summed to generate the joint Q-value which can then be used to calculate the loss like (a centralized form of) DRQN:
\begin{equation}
    \mathcal{L}(\theta)=\mathbb{E}_{<\jh,\ja,r,\jo>\sim\mathcal{D}}\Big[\big(y - \sum_i^\nrA \Qi^\theta(\hi, \ai)\big)^2 \Big] \text{, where\,\, } y=r + \gamma \sum_i^n \max_{\ai'}\Qi^{\theta^-}(\hi',\ai')
    \label{eq:vdn}
\end{equation}
where $\hi'$ is $\hi\ai\oi$ for $\hi$ taken from $\jh$, $\ai$ from $\ja$, and $\oi$ from $\jo$. 

Since the loss uses the sum of the agent's Q-functions, the result is a set of Q-values whose sum approximates the joint Q-function. 
This approximation will be lossless in extreme cases (e.g., full independence) but it may be close or still permit the agents to select the best actions even if the approximation is poor. 
The approach is also scalable since the max used to create the target value $y$ is not over all agent actions as in the centralized case ($\jaS$), but done separately for each agent in Equation \ref{eq:vdn}.

A simple version of VDN is given in Algorithm \ref{alg:VDN}. The approach is very similar to standard D(R)QN \citep{DQN, DRQN} as well as multi-agent DTE extensions \citep{Omidshafiei17,Tampuu}. The key difference is the calculation of the loss on Line \ref{alg:vdn:loss}, which approximates the joint Q-value by using the sum of the individual Q-values. 

In more detail, the $\Qi$ are estimated using RNNs for each agent parameterized by $\theta_i$, along with using a target network, $\theta_i^-$, and a replay buffer, $\mathcal{D}$.  First, episodes are generated by interacting with the environment (e.g., $\epsilon$-greedy exploration), which are added to the replay buffer and indexed by episode number and time step ($\mathcal{D}^e(\ts)$). 
This buffer typically has a fixed size and new data replaces old data when the buffer is full. Episodes can then be sampled from the replay buffer (either single episodes or as a minibatch of episodes). In order to have the correct internal state, RNNs are trained sequentially.  Training is done from the beginning of the episode until the end, calculating the internal state and the loss at each step  and updating each $\theta_i$ using gradient descent. The target networks $\theta_i^-$ are updated periodically (every $C$ episodes in the code) by copying the parameters from $\theta$.

\begin{algorithm}
\begin{algorithmic}[1]
\STATE set $\alpha$, $\epsilon$, and $C$ (learning rate, exploration, and target update frequency)
\STATE Initialize network parameters $\thetai$ for each $\Qi$ (denoted $\Qi^\theta$)
\STATE {\bf for all} $i$, $\theta_i^- \gets \theta_i$ 
\STATE $\mathcal{D} \gets \emptyset$
\STATE $e \gets 1$ \hfill \COMMENT{episode index}
\FORALL{episodes}
\STATE  {\bf for all} $\hi \gets \emptyset $ \hfill \COMMENT{initial history is empty}
\FOR{$\ts=1$ to $\hor$}
\STATE {\bf for all} $i$, take $\ai$ at $\hi$ from $\Qi^\theta(\hi,\cdot)$ with exploration (e.g., $\epsilon$-greedy)
\STATE  See joint reward $\rT \ts$, and observations $\joT \ts$\hfill 
\STATE append $\ja,\jo,r$ to $\mathcal{D}^e $
\STATE {\bf for all} $\hi \gets \hi\ai\oi$ \hfill \COMMENT{update RNN state of the network}
\ENDFOR
\STATE sample an episode from $\mathcal{D}$
\FOR{$\ts=1$ to $\hor$}
\STATE {\bf for all} $i$, $\hi \gets \emptyset $
\STATE $\ja, \jo, r \gets \mathcal{D}^e(\ts)$
\STATE {\bf for all} $i$, $\hi' \gets \hi\ai\oi$ 
\STATE $y=r + \gamma\sum_i\max_{\ai'}\Qi^{\theta^-}(\hi',\ai')$
\STATE  {\bf for all} $i$, do gradient descent on $\thetai$ with learning rate $\alpha$ and loss  $ \big(y - \sum_i \Qi^{\theta}(\hi, \ai)\big)^2  $ \label{alg:vdn:loss}
\STATE {\bf for all} $i$, $\hi \gets \hi'$
\ENDFOR
\IF{$e \mod C = 0$} 
   \STATE {\bf for all} $i$, $\theta_i^- \gets \theta_i$ 
 \ENDIF
 \STATE $e \gets e+1$
\ENDFOR
\RETURN all $\Qi$ 
\end{algorithmic} 
\caption{A version of value decomposition networks (VDN) (finite-horizon)}
\label{alg:VDN}
\end{algorithm}

While the algorithm is presented for the finite-horizon case (for simplicity), it can easily extended to the episodic and infinite-horizon cases by including terminal states or removing the loop over episodes. 

After training, each agent keeps its own recurrent network, which can output the Q-values for that agent. The agent can then select an action by argmaxing over those Q-values: $\poli(\hi)=\argmax_{\ai}\Qi(\hi,\ai)$. 

VDN is often used a baseline but by using a simple sum to combine Q-functions, it can perform poorly compared to more sophisticated methods.

\subsection{QMIX}

QMIX \citep{QMIX,QMIXJMLR} extends the value factorization idea of VDN to allow more general decompositions. In particular, instead of assuming the joint Q-function, $Q(\jh, \ja)$, factors into a sum of local Q-values, $\Qi(\hi,\ai)$, QMIX assumes the joint Q-function is a monotonic function of the individual Q-functions. The sum used in VDN is also a monotonic function but QMIX allows more general (possibly nonlinear) monotonic functions to be learned. 

Specifically, QMIX assumes the following approximate factorization of the Q-function:
\begin{equation}
\jQ(\jh, \ja) \approx f_{mono}(\Qi(\aoHistA 1,\aA 1),\ldots, Q_n(\aoHistA n,\aA n))
\end{equation}

This monotonic assumption means the argmax over the local Q-functions is also an argmax over the joint Q-function, 
This property allows agents to choose actions from their local Q-functions instead of having to choose them from a joint Q-function---ensuring the choice would be the same in both cases.   
Like in VDN, it also makes the training more efficient since calculating the argmax in the update Equation \ref{eq:qmix} is linear in the number of agents rather than exponential.

\cite{QTRAN} formalized this property as the Individual-Global-Max (IGM) principal. IGM states that the argmax over the joint Q-function is the same as argmaxing over each agent's individual Q-function. The definition below is for a particular history but it should ideally hold for all histories (or at least the ones visited by the policy). 
\begin{definition}[Individual-Global-Max (IGM) \cite{QTRAN}] 
For a joint action-value function $\jQ(\jh, \ja)$, where $\jh=\langle \aoHistA 1,\ldots, \aoHistA n \rangle$ is a joint action-observation history, if there exist individual 
functions [$\Qi$], such that:
    \begin{equation}
  \argmax_{\ja} \jQ(\jh, \ja) = \left( \begin{array}{ll}
      \argmax_{\aA 1}Q_1(\aoHistA 1,\aA 1)\\
          \hfill \vdots  \hfill\\
      \argmax_{\aA n}Q_n(\aoHistA n,\aA n)
    \end{array}
  \right),
\end{equation}
then,  [$\Qi$] satisfy IGM for $Q$ at $\jh$.
  \label{def:IGM}
\end{definition}

\begin{figure}
\begin{centering}

\includegraphics[scale=0.7]{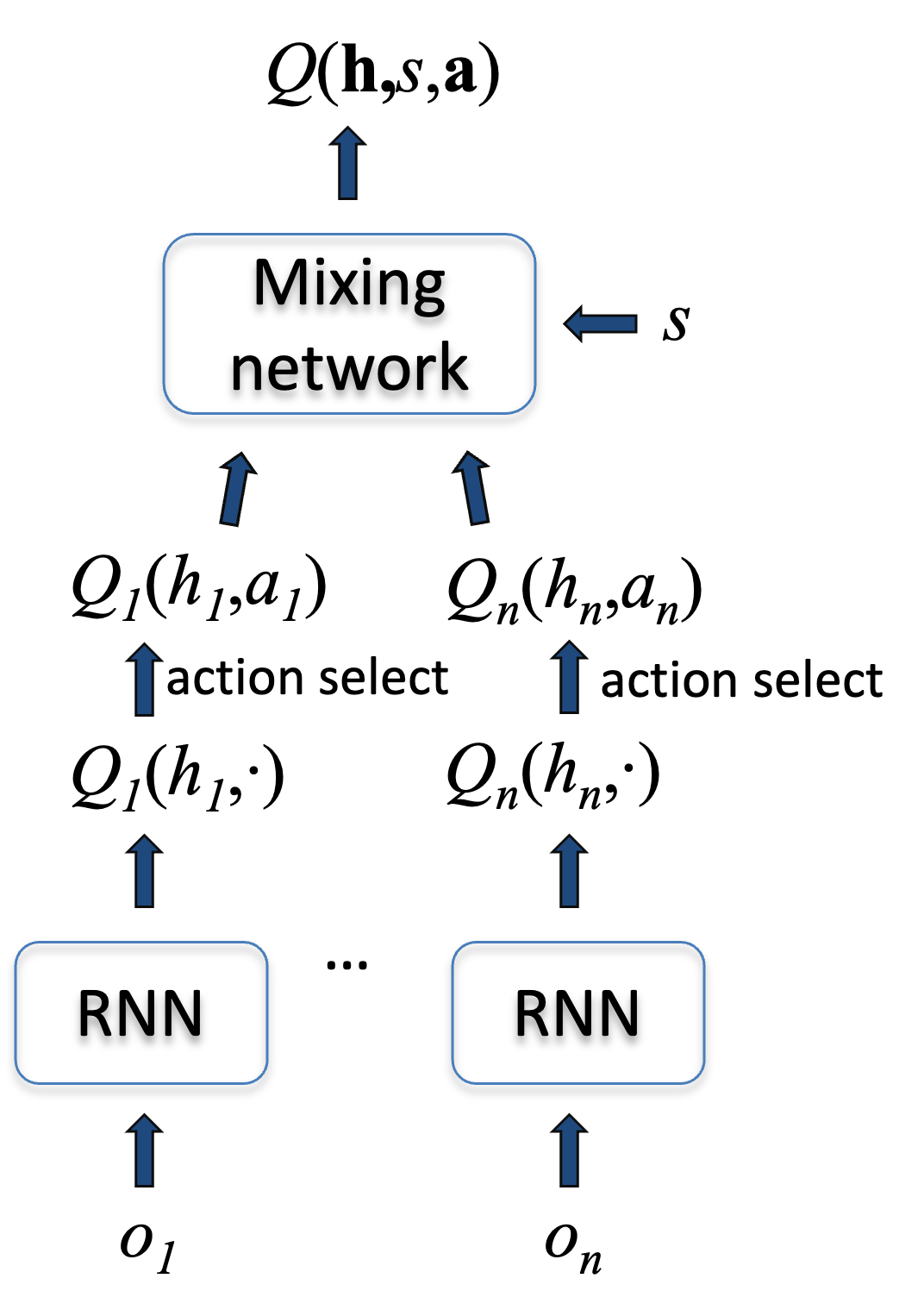}

\end{centering}
\caption{QMIX diagram}

\label{fig:QMIX}
\end{figure}

A simplified version of the QMIX architecture is shown in Figure \ref{fig:QMIX}. Like VDN, each agent has an RNN that takes in the current observation and can output Q-values for the updated history over all actions.\footnote{QMIX typically also inputs the previous action but this is omitted here for clarity and consistency.} A particular action is then chosen (e.g., using $\epsilon$-greedy action selection) using each $\Qi$ and each $\Qi(\hi,\ai)$ is fed into the mixing network. The mixing network is made to be monotonic by restricting the weights (but not the bias terms) to be non-negative. 
The mixing network can also receive the state to potentially improve performance. In practice, the state is given as input to hypernetworks \citep{hypernetworks} which generate the weights for the layers of the mixing network. 
Because the output depends on the state, it is $\jQ(\jh,s,\ja)$, and is often called $\jQ_{tot}$.

The network is trained end-to-end, like VDN by calculating a loss between $\jQ(\jh,s,\ja)$ and the joint return. 
The corresponding loss is:
\begin{multline}
    \mathcal{L}(\theta)=\mathbb{E}_{<\jh,s,\ja,r,\jo,s'>\sim\mathcal{D}}\Big[\big(y - \jQ^\theta(\jh,s,\ja)\big)^2 \Big],  \text{where\, } y=r + \gamma  \jQ^{\theta^-}(\jh',s',\tilde \ja'),   \\\text{ and\, }  \tilde \ja'= \langle \argmax_{\aA 1'} Q_1(\aoHistA 1',\aA 1'),\ldots, \argmax_{\aA n'} Q_n(\aoHistA n',\aA n') \rangle
    \label{eq:qmix}
\end{multline}
where the value on the next time step, $\jQ^{\theta^-}(\jh',s',\tilde \ja')$, is gotten by inputting the argmax over each agent's local Q-value using $\jh'=\langle \aoHistA 1', \ldots, \aoHistA \nrA' \rangle$ as well as the corresponding state $s'$ that is sampled from the buffer at  $\jh'$.

The mixing network is only needed during training. Agents can retain their RNNs (which are just DRQNs) for selecting actions during decentralized execution just like VDN. 

Again, such a factorization may not be possible in all problems so QMIX may not be able to approximate the true joint Q-function in all problems accurately. In particular, as discussed in the paper, this factorization should fail when agents' action choices depend on (at least some) other agents' actions at the same time step. 

\paragraph{Weighted QMIX} extends QMIX in an attempt to improve its expressiveness \citep{WeightedQMIX}.
The main idea is to weigh the actions at different histories differently so the Q-values for some actions can be accurately represented but other action values can be less accurate. 
 For instance, if you knew the optimal policy, you could set the weight highest for the optimal action in each history.  You don't need to accurately represent the other action values but you do need to make sure they are lower than the optimal action (so the policy will choose the optimal action!). 

Specifically, weighted QMIX uses weights $w(\jh,\ja) \in  (0,1]$ to adjust the importance of joint history-action pairs in the loss (as formalized below).\footnote{The paper discusses much of the method and theory in terms of the fully observable case but we consider the partially observable case here. In the partially observable case, the weights can take the history or the history and the state: $w(h,s,a)$. } 
It turns out that there always exists some weighing that will allow the optimal policy to be recovered by independently argmaxing over each agent's $\Qi$. 
An idealized form of the algorithm could also converge in the fully observable case but this proof doesn't extend to the partially observable case or when approximations are used (i.e., the deep RL case).

The method has two main components, a QMIX network where the output is weighted, and an approximation of the optimal Q-value called $\hat \jQ^*$.
The first network takes the output of QMIX (as seen in Figure \ref{fig:QMIX}) and weighs it as seen in Equation \ref{eq:wqmix}. For easier differentiation with the other Q network, we call the output of the first (QMIX-style) network $\jQ_{tot}$. This network is exactly the same as the one in QMIX but the output is weighted in the loss to prioritize different actions at different histories. 
 The $\hat \jQ^*$ network is similar to the one in QMIX but it does not limit the weights to be non-negative and doesn't use a hypernetwork (just directly inputting the state into the mixing network). These networks don't share parameters and are trained using the losses given in Equation  \ref{eq:wqmix}. The same target ($y$) is used in both cases, which now uses the unconstrained output  $\hat \jQ^*$. The argmax is done the same way as in QMIX, where the actions are chosen using each agent's $\Qi$. As a result, the argmax is tractable but a potentially more accurate value of those actions can be provided from  $\hat \jQ^*$ rather than $\jQ_{tot}$. 
 
 In particular, the losses uses are:
\begin{equation}
\begin{split}
 &\mathcal{L}(\theta_{cent})=\mathbb{E}_{<\jh,s,\ja,r,\jo,s'>\sim\mathcal{D}}\Big[\big(y - \hat \jQ^*(\jh,s,\ja)\big)^2 \Big], \hfill \\
   &  \mathcal{L}(\theta_{tot})=\mathbb{E}_{<\jh,s,\ja,r,\jo,s'>\sim\mathcal{D}}\Big[w(\jh,\ja)\big(y - \jQ_{tot}(\jh,s,\ja)\big)^2 \Big],  \\
    & \text{where\, } y=r + \gamma  \hat \jQ^*(\jh',s',\tilde \ja'),  \text{ and\, }  \tilde \ja'= \langle \argmax_{\aA 1'} Q_1(\aoHistA 1',\aA 1'),\ldots, \argmax_{\aA n'} Q_n(\aoHistA n',\aA n') \rangle\\
    \label{eq:wqmix}
\end{split}
\end{equation}

Two different weighting functions are considered. Centrally-Weighted QMIX (CW) uses:
  $$ w(\jh,\ja)= \bigg\{
 \begin{array}{ll}
      1 \quad \text{if \,} y > \hat \jQ^*(\jh,s,\tilde \ja^*) \text{\, or \,} \ja=\tilde \ja^* \\
      \alpha \quad  \text{\, otherwise}\\
    \end{array}$$
which is an approximation of the optimal action $\tilde \ja^*$ using the individual argmax as in Equation \ref{eq:wqmix} and an approximation of the value function, $\hat Q^*$, using the unconstrained network. The weight is set to $1$ when the action is already known to be optimal ($\ja=\tilde \ja^*$) or when the action has a higher value than the current best action.   
Optimistically-Weighted QMIX (OW) uses:
        $$ w(\jh,\ja)= \bigg\{
 \begin{array}{ll}
      1 \quad \text{if \,} y>\jQ_{tot}(\jh,s,\ja)  \\
      \alpha \quad  \text{\, otherwise}\\
    \end{array}$$
which just sets the weight to $1$ when the current estimate using $\jQ_{tot}$ is less than the target value, which uses $\hat \jQ^*$. This inequality suggests error in $\jQ_{tot}$ and $\ja$ could (optimistically) be optimal at $\jh$. 
$\alpha$ is a hyperparamter that can be set during training. 

 After training, the $\Qi$ from the constrained network (outputting $\jQ_{tot}$) can be used for decentralized action selection (just like in QMIX). Therefore, $\hat \jQ^*$ is only used to help guide the learning so better (hopefully, optimal) actions will have higher values in each agent's $\Qi$. 
While weighted QMIX can outperform QMIX in some cases, it isn't as widely used as more recent methods (e.g., QPLEX \citep{QPLEX} and MAPPO \citep{MAPPO}). 

\subsection{QTRAN}

QTRAN \citep{QTRAN} provides a different way to factor the joint Q-function into individual Q-functions. The idea generalizes VDN but the intuition is that the sum does not have to equal the true Q-function but some \emph{transformed} Q-function $\jQ'$. A separate V term\footnote{While the paper calls this a state value, the value takes histories as input, not states.} can then be used to account for the error between the transformed $\jQ'$ and  the true $\jQ$. The approach cannot represent all IGM-able functions but it is more general than VDN.\footnote{While some have taken the text of the paper to show that QTRAN can represent all IGM-able functions, the proofs only show that the method satisfies IGM (not that any IGM-able function can be represented by the method).} QPLEX (which I talk about below) is more general and the relationship to QMIX and weighted QMIX is unclear. 

Specifically, QTRAN uses Equation \ref{eq:qtran} as the basis for their architecture and losses. 

    \begin{equation}
\sum_{i} \Qi(\hi,\ai) - \jQ(\jh,\ja)+\jV(\jh)= \bigg\{
 \begin{array}{ll}
      0 \quad\quad \ja=\tilde \ja\\
      \ge 0 \quad \ja\ne\tilde \ja\\
    \end{array}
    \label{eq:qtran}
\end{equation}
where $$\jV(\jh)=\max_{\ja}\jQ(\jh,\ja) -\sum_{i} \Qi(\hi,\tilde \ai),$$  
 $$\tilde \ai=  \argmax_{\ai} \Qi(\hi,\ai), \quad \text{and} \quad \tilde \ja= \langle \tilde {\aA 1}, \ldots,  \tilde {\aA n} \rangle$$
Here,  $\jV(\jh)$ has a particular form where it is the difference between the (centralized) max over the joint Q-function, $\jQ$, and the sum of the maxes over the individual Q-functions, $\Qi$. Equation \ref{eq:qtran} is constrained to be $0$ when the actions are the argmax over the individual Q-functions, $\tilde \ja$, and greater than or equal to 0 for other actions. If Equation \ref{eq:qtran} holds, the $\Qi$ satisfy IGM for the $\jQ$ but it isn't clear how general it is. That is, unlike QPLEX, the proof does not show that all functions that satisfy IGM can be represented in this form. 

Instead of enforcing a network structure to get IGM like VDN and QMIX, QTRAN does so by using losses based on Equation \ref{eq:qtran}. The architecture outputs $\jQ'(\jh,\ja)=\sum_i  \Qi(\hi, \ai)$, like VDN, as well as an unconstrained $\jQ(\jh,\ja)$ and $\jV(\jh)$. 
As shown below, there is the standard TD loss for learning the true joint Q-function, $\jQ$, a loss for learning the transformed Q-function, $\jQ'$, and value offset,  $\jV$, for the given $\jQ$ when the actions are currently the maximizing ones, $\tilde \ja$, in $\mathcal{L}_{opt}(\theta)$ and for other actions in  $\mathcal{L}_{nopt}(\theta)$. The notation $\bar \jQ$ is used to note that  $\jQ$  isn't trained from $\mathcal{L}_{opt}(\theta)$ and  $\mathcal{L}_{nopt}(\theta)$. 
\begin{equation}
\begin{split}
 &   \mathcal{L}_{td}(\theta)=\mathbb{E}_{<\jh,\ja,r,\jo>\sim\mathcal{D}}\Big[\big(y -  \jQ(\jh,\ja)\big)^2 \Big], \\
  &   \mathcal{L}_{opt(\theta})=\mathbb{E}_{<\jh,\ja,r,\jo>\sim\mathcal{D}}\Big[\big(\jQ'(\jh,\tilde \ja) - \bar \jQ(\jh,\tilde \ja)+\jV(\jh)\big)^2 \Big], \\
   &   \mathcal{L}_{nopt}(\theta)=\mathbb{E}_{<\jh,\ja,r,\jo>\sim\mathcal{D}}\Big[\Big(\min\big[\jQ'(\jh, \ja) - \bar \jQ(\jh, \ja)+\jV(\jh),0\big]\Big)^2 \Big], \\
  &  \text{where\, } y=r + \gamma  \jQ_{\theta^-}(\jh',\tilde \ja'),  \text{ and\, }  \tilde \ja'= \langle \argmax_{\aA 1'} Q_1(\aoHistA 1',\aA 1'),\ldots, \argmax_{\aA n'} Q_n(\aoHistA n',\aA n') \rangle\\
    \label{eq:qtranloss}
    \end{split}
\end{equation}

The losses above are used for the standard form of QTRAN called QTRAN-base. An alternate form, called QTRAN-alt, replaces  $\mathcal{L}(\theta_{nopt})$ with the average counterfactual loss below, forcing the value to be 0 for some action rather than relying on the inequality constraint above.  
$$ \mathcal{L}(\theta_{nopt-min})=\mathbb{E}_{<\jh,\ja,r,\jo>\sim\mathcal{D}}\Big[\frac{1}{n}\sum_i^n\Big(\min \jQ'(\jh,\ai, \ja_{-i}) - \bar \jQ(\jh,\ai, \ja_{-i})+\jV(\jh)\Big)^2 \Big], 
$$
Using this condition also satisfies IGM. 

The versions of QTRAN (QTRAN-base and QTRAN-alt) can outperform VDN and QMIX on some domains but they are often outperformed by more recent methods.

\subsection{QPLEX}

QPLEX \citep{QPLEX} further extends the value factorization idea to (provably) include more general decentralized policies. In particular, QPLEX can potentially learn any set of Q-functions that satisfy the IGM principle. 

First, QPLEX extends the IGM principle to an advantage-based case. They
define joint values and advantages from the joint Q-function:
$\jV(\jh)=\max_{\ja} \jQ(\jh,\ja)$, and 
$\jA(\jh,\ja)=\jQ(\jh,\ja)-\jV(\jh)$ as well as 
 individual values and advantages from each agent's individual Q-function:
$\Vi(\hi)=\max_{\ai} \Qi(\hi,\ai)$, and
$\Ai(\hi,\ai)=\Qi(\hi,\ai)-\Vi(\hi)$. 
Note that this notion of advantage is a bit different than the typical notion (such as \citep{DuelingDQN}) since the values and advantages are generated from the Q-values, rather than the other way around. As a result, the advantages will have the property that they will be 0 for optimal actions and negative otherwise. That is, $\Ai(\hi,\ai^*)= 0$ (since $\Qi(\hi,\ai^*)-\max_{\ai}\Qi(\hi,\ai)=0$) and for some non-optimal action $\ai^{\dagger}$, $\Ai(\hi,\ai^{\dagger}) < 0$ (since $\Qi(\hi,\ai^{\dagger})<\max_{\ai}\Qi(\hi,\ai)$). The same holds true for joint advantages. 

Advantage-based IGM extends IGM to the advantage case and states that the argmax over the joint advantage function is the same as argmaxing over each agent's individual advantage function:
\begin{definition}[Advantage-based (IGM) \cite{QPLEX}] 
For a joint action-value function, if there exist individual 
functions [$\Qi$], such that:
    \begin{equation}
  \argmax_{\ja} \jA(\jh, \ja) = \left( \begin{array}{ll}
      \argmax_{\aA 1}A_1(\aoHistA 1,\aA 1)\\
           \hfill \vdots  \hfill\\
      \argmax_{\aA n}A_n(\aoHistA n,\aA n)
    \end{array}
  \right).
\end{equation}
where $\jA(\jh, \ja)$ and $\Ai$ are defined above, then [$\Qi$] satisfy advantage-based IGM for $\jQ$ at $\jh$.
\end{definition}
That is, advantage-based IGM is equivalent to the Q-based IGM in Definition \ref{def:IGM}. 

\begin{figure}
\begin{centering}

\includegraphics[scale=0.7]{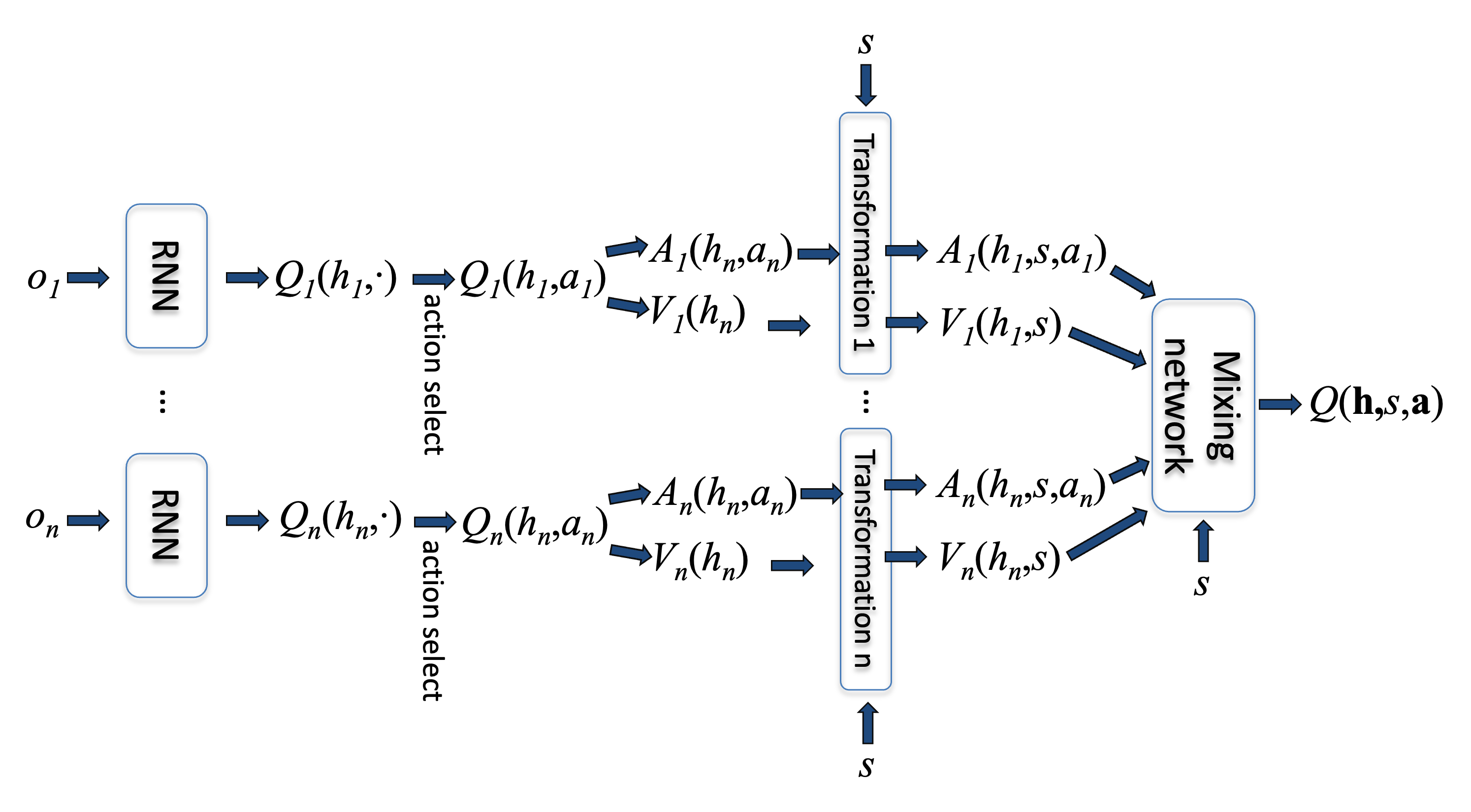}

\end{centering}
\caption{QPLEX diagram}

\label{fig:QPLEX}
\end{figure}

QPLEX uses advantage-based IGM to extend QMIX and QTRAN to represent the full IGM function class. 
The architecture is shown in Figure \ref{fig:QPLEX}. 
Like QMIX (and VDN and QTRAN), QPLEX first takes as input each agent's observation $\oi$ (and previous action $\ai^{t-1}$ but that is not included here) and outputs the individual Q-values, $\Qi(\hi,\ai)$, for all $\ai$ using a DRQN-style network. After an action is selected (e.g., using $\epsilon$-greedy exploration), the value and advantage are extracted from the Q-values for each agent as $\Vi(\hi)=\max_{\ai} \Qi(\hi,\ai)$ and $\Ai(\hi,\ai)=\Qi(\hi,\ai) - \Vi(\hi)$. 
Next, the transformation networks generate $\Vi(\hi,s)$ and $\Ai(\hi,s,\ai)$ for each agent with the given $\Vi(\hi)$ and $\Ai(\hi,\ai)$ along with  $s$ as $\Vi(\hi,s)=w_i(s)\Vi(\hi)+b_i(s)$ and $\Ai(\hi,s,\ai)=w_i(s)\Ai(\hi,\ai)$. Finally, the joint Q-value is output based on the output of each agent's transformation network and the state as $\jQ(\jh,s,\ja) = \sum_i \Vi(\hi,s)+\lambda_i(s,\ja) \Ai(\hi,s,\ai)$.\footnote{Attention is used to train the $\lambda$ weights more efficiently.} 
While there aren't constraints on V due to IGM (because it doesn't include the actions), QPLEX uses a sum to combine the local V's. 
All the weights, $w_i$ and $\lambda_i$ (but not necessarily the biases $b_i$) are positive to maintain monotonicity (like QMIX) and thus satisfy the advantage IGM principle.
The architecture is trained using  the RL loss, just like QMIX in Equation \ref{eq:qmix}.

While positive weights are used (like in QMIX), there are separate weights for $A$ and $V$, and the weights for V don't depend on the action. 
As a result, QPLEX is more general than previous methods with the capacity to represent all Q-functions that satisfy IGM. 
QPLEX also performs well, often outperforming other value factorization methods and it is currently one of the best-performing CTDE methods.

\subsection{The use of state in factorization methods}
\label{sec:vff:state}
Many of the current methods use inconsistent notation about the inclusion of state (except QMIX). As a result, the theory is often developed without considering the state input. 
It turns out that, unlike the policy gradient case discussed in Section \ref{sec:ctde:pg:states}, the using the state doesn't introduce bias and is thus theoretically sound \citep{Marchesini24}. The intuition is that the state is additional information in these cases. It doesn't replace the history, but potentially augments it (similar to a history-state critic in Section \ref{sec:ctde:pg:states}). Furthermore, actions are not chosen based on the state but only the local, history-dependent Q-values, $\Qi$. 
Nevertheless, in implementations of algorithms such as QPLEX and weighted QMIX, the weights only take the state as input and not the history. This replacement of history with state in these contexts limits the representational power of the weights in partially observable problems. As a result, using the state as input in QPLEX (instead of the joint history) prevents it from being able to represent the full IGM class of functions. It still may be beneficial to use the state instead of the joint history but exploring what additional information to use and in what way is an interesting open question.

\section{Centralized critic methods}
\label{sec:ctde:pg}

Concurrently, Multi-Agent Deep Deterministic Policy Gradient (MADDPG)~\citep{MADDPG} and COunterfactual Multi-Agent policy gradient (COMA)~\citep{COMA} popularized the use of centralized critics in MARL. 
I will first discuss the general class of algorithms that MADDPG and COMA represent---multi-agent actor-critic methods with centralized critics---and then give the details of MADDPG and COMA. I will then describe the very popular PPO-based extension, MAPPO~\citep{MAPPO}. 
Finally, I will discuss the (incorrect) use of state in centralized critics as well as the theoretical and practical tradeoffs in the various types of critics.

\subsection{Preliminaries}

\begin{figure}
\begin{centering}
\hfill
\subfigure[]{
\includegraphics[scale=0.25]{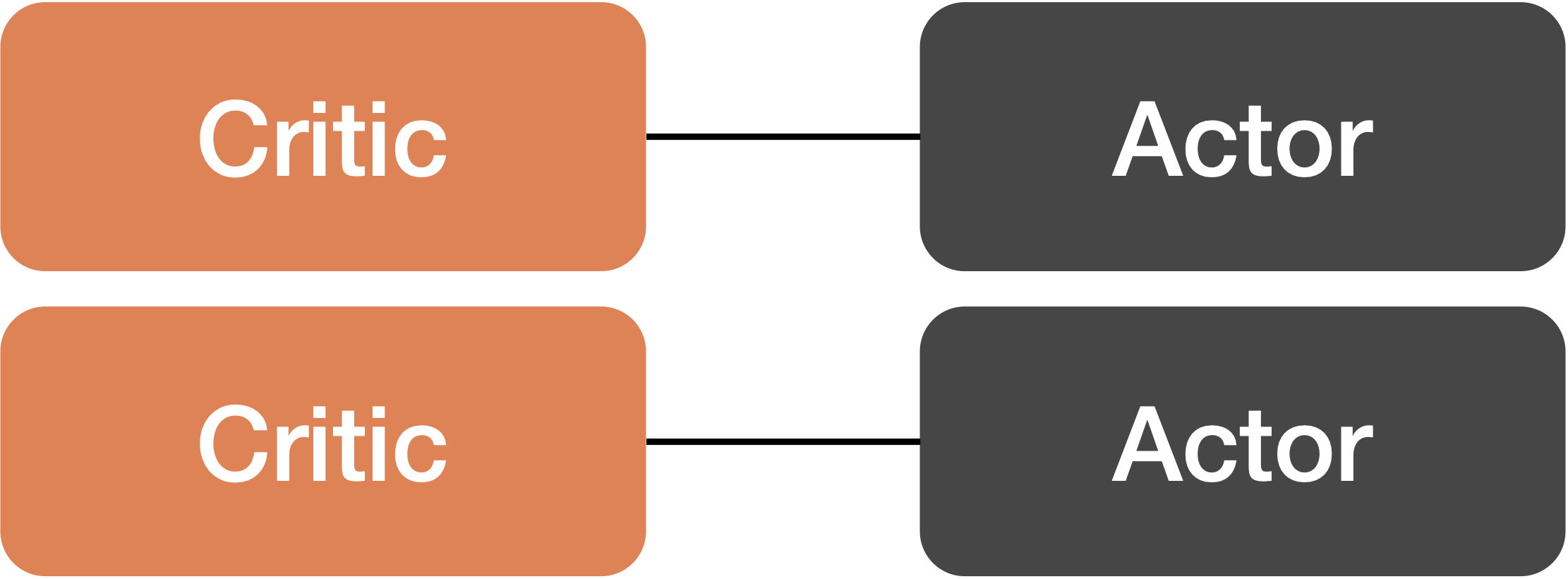} \label{fig:IAC}
}
\hfill
\subfigure[]{
\includegraphics[scale=0.25]{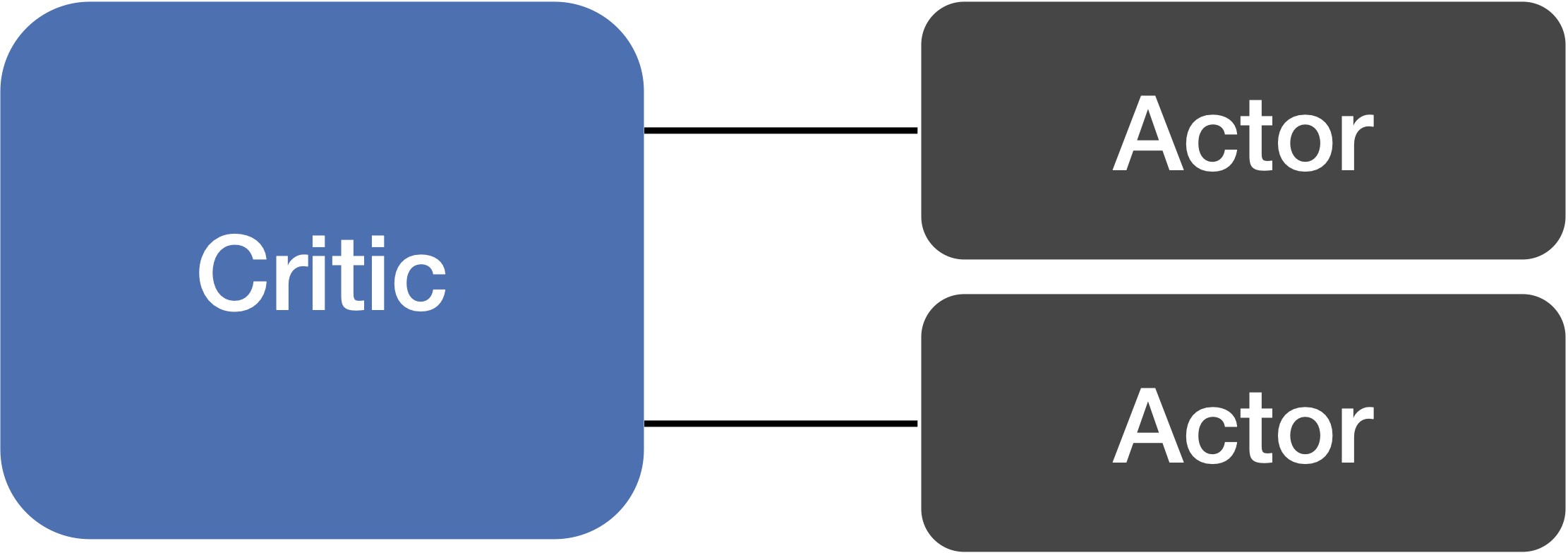} \label{fig:IACC}
}
\hfill
\end{centering}
\caption{Decentralized critics (a) vs.~a centralized critic (b). }
\end{figure}

Single-agent policy gradient and actor-critic methods have also been extended to the Dec-POMDP case. In the multi-agent case, there is one actor per agent and either one critic per agent (as shown in \ref{fig:IAC} and discussed in \citep{DTE}) or a shared, centralized critic (as shown in \ref{fig:IACC} and discussed below). The centralized critic can be used during CTDE but then each agent can act in a decentralized manner by using its actor. The basic motivation is to leverage the centralized information that is available during training and potentially combat nonstationarity that can happen in decentralized training. All the policy gradient approaches can generally scale to larger action spaces and have stronger (local) convergence guarantees than the value-based counterparts above. 

Policy gradient methods use continuous-action or stochastic policies. 
Unless stated otherwise, I will assume a stochastic policy for each agent parameterized by $\ppi$, where $\poli^{\ppi}(\ai| \hi)$ represents the probability agent $i$ will choose action $\ai$ given the history $\hi$ and parameters, $\ppi$, $\Pr(\ai| \hi,\ppi)$. This is in contrast to the value-based methods in Section \ref{sec:ctde:vff} where deterministic policies were generated based on the learned value functions. 
Like the value-based methods, the policy gradient methods also assume algorithms receive agent actions, $\ja$, the resulting observations, $\jo$, and the joint reward, $r$ at each time step.

\subsection{A basic centralized critic approach}

In the simplest form, we can consider a class of centralized critic methods where the joint history critic, which we denote as $\jQmodel(\jh, \ja)$, 
 estimates $\jQpol(\jh, \ja)$ with parameters $\valparam$. $\jQmodel(\jh, \ja)$ is then used to update each decentralized policy $\poli$. 
We call this approach independent actor with centralized critic (IACC) and it is described in Algorithm \ref{alg:IACC}. 
Learning is over a set of (on-policy) episodes and the histories are initialized to be empty. Agent $i$'s history  at time step $\ts$ is denoted $\hiT \ts$, while the set of histories for all agents is denoted $\jh_{\ts}$. All agents choose an initial action from their policies $\aiT 0 \sim \poli(\ai| \hiT 0)$ and then the learning takes place over a fixed horizon of $\hor$. At each step, the agents take the corresponding current action $\jaT \ts$, receive the joint reward $\rT \ts$ and see observations $\joT \ts$. The histories for the next step are updated to include the action taken and observation seen, $\hiT {\ts} \aiT \ts \oiT \ts$, and actions are sampled for the next step, $\aiT {\ts+1} \sim \poli(\ai| \hiT {\ts+1})$. Then, the TD error can be computed with the current Q-values, and then is used to update the actors and critic. Each agent's actor is updated using the TD error from the joint Q-function (moving in the direction that improves the joint Q-value). The centralized critic is updated using a standard on-policy gradient update. The action for the next step, $\ts+1$, becomes the action for the current step, $\ts$, and the process continues until the end of the horizon and for each episode. 
The algorithm is written for the finite-horizon case but it  can be adapted to the episodic or infinite-horizon case by including terminal states or removing the loop over episodes. Note that the histories make the value functions based on a particular time step as the history length provides the time step.

While a sample is used in the algorithm, the full gradient associated with each actor in IACC is represented as:
\begin{equation}
     \nablai J = \mathbb{E}_{<\jh, \ja>\sim\mathcal{D}} \left[ \jQpol(\jh, \ja) \nablai \log \poli(\ai|\hi) \right] \,,
 \label{eq:gradient:qhs}\footnote{For simplicity and to match common implementations, we ignore discounting but the true gradient would include it \citep{nota2020policy}.}
\end{equation}
where the approximation of the joint Q-function, $\jQmodel(\jh, \ja)$, is used to update the policy parameters of agent $i$'s actor, $\poli(\ai| \hi)$. Histories (i.e., observations) and actions are sampled according to the on-policy (discounted) visitation probability. 

The objective, $J$, is to maximize expected (discounted) return starting from the initial state distribution $\bO$ as discussed in Section \ref{sec:decpomdp}. 

The critic is updated using the on-policy loss: 
 $$   \mathcal{L}(\theta)=\mathbb{E}_{<\jh, \ja, r, \jh'>\sim\mathcal{D}}\Big[\big(y - \jQmodel(\jh, \ja)\big)^2 \Big] \text{, where\,\, } y=r + \gamma  \jQmodel(\jh',\ja')$$

\begin{algorithm}
\caption{Independent Actor Centralized Critic (IACC) (finite-horizon)}
\begin{algorithmic}[1]
\label{alg:IACC}
\STATE Initialize individual actor models $\poli(\ai| \hi)$, parameterized by $\ppi$
\STATE Initialize centralized critic model $ \jQmodel(\jh,\ja)$, parameterized by  $\valparam$

\FORALL{episodes}
	\STATE  $\hiT 0 \gets \emptyset$ \hfill  \COMMENT{Empty initial history}
	\STATE Denote $\jh_{t}$ as  $\langle \aoHistAT 1 0,\ldots,\aoHistAT n 0\rangle$ \hfill \COMMENT{Notation for joint variables}
	\STATE {\bf for all} $i$, choose $\aiT 0$ at $\hiT 0$ from $\poli(\ai| \hiT 0)$
	\STATE Store $\ja_{\ts}$ as  $\langle \aAT 1 0,\ldots,\aAT n 0\rangle$
	\FOR{$\ts=0$ to $\hor-1$}
		\STATE Take joint action $\ja_{\ts}$, see joint reward $\rT \ts$, and observations $\joT \ts$\hfill 
		\STATE {\bf for all} $i$, $\hiT {\ts+1} \gets \hiT {\ts} \aiT \ts \oiT \ts$\hfill \COMMENT{Append new action and obs to previous history}
		\STATE {\bf for all} $i$, choose $\aiT {\ts+1}$ at $\hiT {\ts+1}$ from $\poli(\ai| \hiT {\ts+1})$
		\STATE Store $\ja_{\ts+1}$ as  $\langle \aAT 1 {\ts+1},\ldots,\aAT n {\ts+1}\rangle$
	 \STATE $\delta_t \gets r_t + \gamma \jQmodel(\jh_{t+1},\ja_{t+1}) - \jQmodel(\jh_t,\ja_t)$ \hfill \COMMENT{Compute centralized TD error}
	 
        \STATE Compute critic gradient estimate: $ \delta_{t} \nablasub \valparam \jQmodel(\jh_t,\ja_t)$ \label{alg:IACC:critic}
        \STATE Update critic parameters $\valparam$ using gradient estimate (e.g., $\valparam \gets \valparam + \beta  \delta_{t} \nablasub \valparam \jQmodel(\jh_t,\ja_t)$ for learning rate $\beta$)
        
            \FOR{each agent $i$} \label{alg:IACC:actor}
       		 \STATE Compute actor gradient estimate: $ \gamma^t \jQmodel(\jh_t,\ja_t) \nablai \log\poli(a_{i,t}| h_{i,t})$ 
                \STATE Update actor parameters $\ppi$ using gradient estimate (e.g., $\ppi \gets \ppi + \alpha \gamma^t \jQmodel(\jh,\ja) \nablai\log\poli(a_{i,t}| h_{i,t})$ for learning rate $\alpha$)
    \ENDFOR
	\ENDFOR
\ENDFOR
\end{algorithmic}
\end{algorithm}

We can extend this algorithm to the more commonly used advantage actor-critic (A2C) case, as shown in Algorithm \ref{alg:IA2CC}. The advantage is defined as $\jA(\jh,\ja)=\jQ(\jh,\ja)-\jV(\jh)$, representing the difference between the Q-value for a given action and the V-value at that history.  
Because $\jV(\jh)$ doesn't depend on the action, it is called a \emph{baseline} and becomes a constant from the perspective of the policy gradient. As a result using $\jQ(\jh,\ja)-\jV(\jh)$ in place of $\jQ(\jh,\ja)$ doesn't change the convergence properties of the method. 
Nevertheless, using a baseline (e.g., subtracting $\jV(\jh)$) can reduce the variance of the estimate and improve performance in practice. In the algorithm, $\jA(\jh,\ja)$ isn't explicitly stored since the TD error is used as an approximation for the advantage---$\jQmodel(\jh,\ja)$ is approximated with $r_t + \gamma \jVmodel(\jh_{t+1})$, which is true in expectation.  $\delta_t$ can be thought of as a sample of  $\jQmodel(\jh_t,\ja_t)-\jVmodel(\jh_t) = \jAmodel(\jh_t,\ja_t)$.

\begin{algorithm}
\caption{Independent Advantage Actor Centralized Critic (IA2CC) (finite-horizon)}
\begin{algorithmic}[1]
\label{alg:IA2CC}
\STATE Initialize individual actor models $\poli(\ai| \hi)$, parameterized by $\ppi$
\STATE Initialize  centralized critic model $\hat \jV(\jh)$, parameterized by  $\valparam$

\FORALL{episodes}
	\STATE {\bf for all} $i$, $\hiT 0 \gets \emptyset$ \hfill  \COMMENT{Empty initial history}
	\STATE Denote $\jh_{t}$ as  $\langle \aoHistAT 1 0,\ldots,\aoHistAT n 0\rangle$ \hfill \COMMENT{Notation for joint variables}
	\STATE {\bf for all} $i$, choose $\aiT 0$ at $\hiT 0$ from $\poli(\ai| \hiT 0)$
	\STATE Store $\ja_{\ts}$ as  $\langle \aAT 1 0,\ldots,\aAT n 0\rangle$
	\FOR{$\ts=0$ to $\hor-1$}
		\STATE Take joint action $\ja_{\ts}$, see joint reward $\rT \ts$, and observations $\joT \ts$\hfill 
		\STATE {\bf for all} $i$, $\hiT {\ts+1} \gets \hiT {\ts} \aiT \ts \oiT \ts$\hfill \COMMENT{Append new action and obs to previous history}
		\STATE {\bf for all} $i$, choose $\aiT {\ts+1}$ at $\hiT {\ts+1}$ from $\poli(\ai| \hiT {\ts+1})$
		\STATE Store $\ja_{\ts+1}$ as  $\langle \aAT 1 {\ts+1},\ldots,\aAT n {\ts+1}\rangle$
	 \STATE  $\delta_t \gets r_t + \gamma \jVmodel(\jh_{t+1}) - \jVmodel(\jh_t)$ \hfill \COMMENT{Compute centralized advantage estimates (TD error)}
	 
        \STATE Compute critic gradient estimate: $ \delta_{t} \nabla \jVmodel(\jh_t)$\label{alg:IA2CC:critic}
        \STATE Update critic parameters $\valparam$ using gradient estimate (e.g., $\valparam \gets \valparam + \beta \gamma \delta_{t} \nabla \jVmodel(\jh_t))$
        
           \FOR{each agent $i$} \label{alg:IA2CC:actor}
        \STATE Compute actor gradient estimate: $ \gamma^t \delta_t \nabla\log\poli(a_{i,t}| h_{i,t})$ \label{alg:IA2CC:actorgrad}
        \STATE Update actor parameters $\ppi$ using gradient estimate (e.g., $\ppi \gets \ppi + \alpha \gamma^t \delta_{t} \nabla\log\poli(a_{i,t}| h_{i,t})$)
    \ENDFOR
 	\ENDFOR
\ENDFOR
\end{algorithmic}
\end{algorithm}

While the critic is called \emph{centralized} because it uses centralized information, it estimates the \emph{joint} value function of the decentralized policies. That is, the centralized critic estimates the value of the current set of decentralized policies (not a centralized policy) \citep{JAIR23}. This is the correct thing to do since the critic's job is to evaluate the policies, which are decentralized in this case. 

With appropriate assumptions (e.g., on exploration, learning rate, and function approximation), IACC (and IA2CC) will converge to a local optimum \citep{Peshkin00UAI,JAIR23,Lyu24}. These results put policy gradient methods on solid theoretical ground and I discuss details of the various methods as well as some theoretical shortcomings. 

This idea of learning a centralized critic to update decentralized actors is very general. It can be (and has been) used with different types of critics, actors, and updates. I present some of the most popular methods below.

\subsection{MADDPG} 
\label{sec:ctde:pg:maddpg}
Multi-Agent DDPG (MADDPG) \citep{MADDPG} considers the more general case of possibly competitive agents as well as continuous actions. To deal with the different reward functions of different agents in the competitive case, separate centralized critics are learned for each agent. 
As we are only concerned with the cooperative case, we assume a single shared critic among agents, do not have to learn policy models of the other agents (since these are assumed to be accessible in a cooperative CTDE setting), and do not consider ensembles of other agent policies to improve robustness. MADDPG is also an off-policy method, unlike the previous algorithms discussed, so it makes use of replay buffer similar to DQN-based approaches. Nevertheless, MADDPG for the cooperative case is very similar to Algorithm \ref{alg:IACC} with the main changes noted below.

To deal with continuous actions, MADDPG extends the continuous action single-agent deep actor-critic method DDPG \citep{DDPG} to the multi-agent case. 
I denote the deterministic continuous-action policies for each agent as $\mu_i$. 
In the cooperative case, MADDPG uses the following policy gradient:
\begin{equation}
   \nablai J \doteq (1-\gamma) \Exp_{x, \ja\sim\rho(x,\ja)} \left[ \nablai \mu_i(\oi) \nablasub \ja \jQpol(x, \ja) \mid_{a_i=\mu_i(\oi)} \right] \,,
 \label{eq:maddpg:actor}
\end{equation}
where $x$ and $\ja$ are sampled according to the (discounted) visitation probability with $x$ discussed below and $\ja=\langle \aA 1 ,\ldots,\aA  n \rangle$. 
Since we can can no longer sum over actions and weigh their value by their probability as is done in the stochastic policy case, we 
have to consider the change in the Q-value function evaluated at the chosen action $a_i$ for the agent \citep{DPG}.

Note that MADDPG considers reactive policies that only map from the last observation to an action (as seen in $\mu_i$). As noted in Section \ref{sec:singleobs}, the policy, $\mu_i$, should choose actions based on histories, $\hi$, rather than single observations, $\oi$.
The Q-value is from a centralized critic so the paper states $x$ ``could consist of the observations of all agents [...] however, we could also include additional state information if available." 
More generally, $x$ should be history representations for each agent and we discuss the use of states in more detail in \ref{sec:ctde:pg:states}. 
These issues are easily fixable and result in the following policy gradient (we can incorporate the state information later):
\begin{equation}
   \nablai J \doteq (1-\gamma) \Exp_{x, \ja\sim\rho(\jh,\ja)} \left[ \nablai \mu_i( \hi) \nablasub \ja \jQpol(\jh, \ja) \mid_{a_i=\mu_i( \hi)} \right] \,.
 \label{eq:maddpg:actor}
\end{equation}

Because MADDPG is off-policy, it maintains a replay buffer like DQN-based approaches along with `target' copies of both the actor and critic networks, $\mu_{\ppi^-}$ and $Q_{\valparam^-}$. The Q-update (for the history-based case) is then:
\begin{equation}
    \mathcal{L}(\theta)=\Exp_{<\jh, \ja, r, \jh'>\sim\mathcal{D}}\Big[\big(y - Q_{\valparam}(\jh, \ja)\big)^2 \Big] \text{, where\,\, } y=r + \gamma Q_{\valparam^-}(\jh',\ja')\mid_{\ai=\mu^-(\hi)\ \forall i \in \agentS} 
    \label{eq:maddpg:critic}
\end{equation}
MADDPG executes a stochastic exploration policy (e.g., by adding Gaussian noise to the current deterministic policy) while estimating (and optimizing) the value of the deterministic policy. This can be seen because while the action, $\ja$, is sampled from the (behavior policy) dataset, the next action, $\ja'$ is selected by the target policy, $\mu^-$. 

Algorithm \ref{alg:IACC} can be updated to incorporate these changes.  Since the continuous-action policy is deterministic, (Guassian) noise is added when selecting actions to aid in exploration. The approach is off-policy, so the experiences are first stored in a replay buffer (like DQN and other value-based methods in Section \ref{sec:ctde:vff}). Similarly, episodes are sampled from the replay buffer and target networks used (again, like DRQN-based approaches) for the actor and the critic. 

MADDPG is no longer widely used but the ideas (such as the centralized critic) have been adopted extensively. 

\subsection{COMA}
\label{sec:ctde:pg:coma}
The main contributions of Counterfactual Multi-Agent Policy Gradients (COMA) were the introduction of the centralized critic along with a counterfactual baseline \citep{COMA}.
As mentioned above, baselines are common in policy gradient methods since they are high variance. The baseline value is subtracted from the Q-value and it can be anything that is not dependent on the agent's action. 

In the case of COMA, the motivation for the baseline is not only variance reduction but also better credit assignment by subtracting off the perceived contribution to the Q-value from the other agents. Specifically, it marginalizes out the agent actions from the Q-function to get an estimate of what the (counterfactual) Q-function would be while holding the other agent actions fixed. The result is an agent-specific advantage function that subtracts the baseline for agent $i$ from the joint Q-function:
$$\Ai(\jh,\ja)=\jQ(\jh,\ja)-\sum_{\ai'} \poli(\ai'| \hi) \jQ(\jh,\ai',\ja_{-i})$$
This baseline no longer depends on agent $i$'s action (in expectation) so it will not bias the gradient. 
Note that the COMA paper uses a state-based critic and advantage, $\jQ(s,\ja)$ and  $\Ai(s,\ja)$, but this is incorrect as I'll discuss in Section \ref{sec:ctde:pg:states}. 

Rather than have separate baseline networks for each agent, there is a single centralized critic that takes in the other agent actions, $\ja_{-i}$, the joint observation, $\jo$, an agent id, $i$, the proposed action $\ai$, and policy probabilities $\poli(\ai'| \hi)$,
and outputs the advantage for that agent $\Ai(\jh,\ja)$ using the equation above. 
This network can also output the joint Q-values for all agent $i$'s actions, $Q(\jh,\cdot,\ja_{-i})$, given the other agent actions, $\ja_{-i}$, the joint observation, $\jo$, and an agent id, $i$. Both of these values are needed for the algorithm. 

Then, the COMA algorithm can be an extension of Algorithm \ref{alg:IACC}.  The critic update can be calculated in the same way (using the Q-values from the network described above) but the agent-specific advantage is incorporated in the actor update. That is, during the actor update (starting on line \ref{alg:IACC:actor}), the agent-advantage $\Ai(\jh,\ja)$ is calculated for the given agent and $\Ai$ is used in the actor gradient estimate instead of $\jQmodel$. 

While COMA has been very influential, it isn't widely used since newer methods tend to outperform it. 

\subsection{MAPPO}
\label{sec:ctde:pg:mappo}
Multi-Agent PPO (MAPPO) extends PPO~\citep{PPO} to the centralized critic MARL case~\citep{MAPPO}.
The motivation behind PPO is to adjust the magnitude of the policy update to learn quickly without becoming unstable. 
PPO does this using a simple clipped loss that approximates the more complex trust region update \citep{TRPO}.

Like IA2CC (Algorithm \ref{alg:IA2CC}), MAPPO uses an advantage-based update (but not the agent-specific counterfactual one used in COMA). In particular, instead of the loss being $\gamma^t \delta_t \log\poli(a_{i,t}| h_{i,t})$ as reflected on line \ref{alg:IA2CC:actorgrad}, which is an approximation of $\gamma^t \jA_t \log\poli(a_{i,t}| h_{i,t})$,  the loss becomes:
 \begin{equation}
\mathcal{L}^{MAPPO}_{clip}(\ppi)= \min\Big(r_{\ppi,i}\jA,\text{clip}(r_{\ppi,i},1-\epsilon,1+\epsilon)\jA\Big),
 \label{eq:mappo:actor}
 \end{equation}
 where $r_{\ppi,i} = \frac{\pi_{\ppi}(\ai|\hi)}{\pi_{\polparam_{i,old}}(\ai|\hi)}$ .\footnote{Here, I consider the stochastic (not continuous) action case and for simplicity do not include GAE, mini-batching, or a policy entropy term (which is common in policy gradient methods to improve exploration). Also, the paper uses a local advantage loss, $\Ai$, but this value would not be different than $\jA$ in a general Dec-POMDP if it is updated using  $\jA$ and $r$ since these quantities would be the same for all agents. Finally, policies over single observations rather than histories are used, which as noted before is typically not sufficient for partially observable problems (general Dec-POMDPs) and state-based critics are used, which is incorrect as discussed below.}
 Maximizing $\frac{\pi_{\ppi}(\ai|\hi)}{\pi_{\polparam_{i,old}}(\ai|\hi)}\jA$ seeks to maximally improve the new policy, $\pi_{\ppi}$, compared to the old policy, $\pi_{\polparam_{i,old}}$, by reweighing the advantage using an importance sampling ratio (considering the advantage was calculated using the old policy). 
 It can be difficult to estimate this value from a small number of samples and it may result in too large of an update (resulting in parameters that perform worse). As a result, PPO also considered a term using a clipped ratio, limiting $r_{\ppi,i}$ so it can't be too far above or below 1 (which would be the value when the new policy is equal to the old one). By minimizing over the unclipped and clipped values, PPO limits changes in the policy. Note the ratio is always positive but the advantage could be positive or negative. When the advantage is positive, the loss will be clipped if the ratio is too large, limiting how much more likely the action can be in the policy. Similarly, when the advantage is negative, the loss will be clipped when the ratio is near 0, limiting how much less likely the action can be in the policy. 
Like in IA2CC, $\jA$ can still be approximated as $r_t + \gamma \jVmodel(\jh_{t+1}) - \jVmodel(\jh_t)$.

The critic loss for MAPPO also includes clipping and is given by:
\begin{equation}
    \mathcal{L}^{MAPPO}(\valparam)=\max \left[ (\jV(\jh_t)-\hat R_t)^2,\left(\text{clip}(\jV(\jh),\jV_{old}(\jh)-\epsilon ,\jV_{old}(\jh)+\epsilon)-\hat R_t\right)^2 \right]
    \label{eq:mappo:critic}
\end{equation}
where $\hat R$ are Monte Carlo returns starting from the current history (i.e., $\hat R_t=\sum_{j=t}^\hor \gamma r_j$) and $\jV_{old}$ represents the value function from the previous time step but TD can also be used and value clipping does not have to be used.

All agents use the same centralized critic and parameter sharing can also used so all agents share the same actor network.

\paragraph{Independent PPO (IPPO)} is a version of MAPPO where each agent uses its own local advantage rather than the joint advantage. As a result, the IPPO actor loss becomes:
 \begin{equation}
\mathcal{L}^{IPPO}_{clip}(\ppi)= \min\Big(r_{\ppi,i}\Ai,\text{clip}(r_{\ppi,i},1-\epsilon,1+\epsilon)\Ai\Big),
 \label{eq:oppo:actor}
 \end{equation}
 with $\hat \Ai=r_t + \gamma \Vimodel(\hiT {t+1}) - \Vimodel(\hiT t)$. 
The IPPO critic loss becomes: 
\begin{equation}
    \mathcal{L}^{IPPO}(\valparam)=\max \left[ (\Vi(h_{i,t}))-\hat R_t)^2,\left(\text{clip}(\Vi(h_{i,t})),V_{i,old}(h_{i,t}))-\epsilon ,V_{i,old}(h_{i,t}))+\epsilon)-\hat R_t\right)^2 \right]
    \label{eq:ippo:critic}
\end{equation}
where $\hat R$ are Monte Carlo returns starting from the current \emph{local} history, $\Vi(h_{i,t})$. Parameter sharing is used with IPPO so all agents share the actor and critic network, making it a CTDE approach. As a result, there is only one actor and one critic, but it is updated using the data from all agents. \\

MAPPO and IPPO perform very well on the standard benchmark domains. Both algorithms typically perform similarly, but a version of MAPPO where the information in the critic was hand designed to only include features that are relevant to the task could outperform the standard versions of MAPPO and IPPO. This makes sense as the agents did not have to learn what information was necessary, which is difficult in high-dimensional settings. 

\subsection{State-based critics}
\label{sec:ctde:pg:states}

Critics that use state (and not history) are unsound in partially observable environments (i.e., general Dec-POMDPs). 
That is, they are incorrect and could be arbitrarily bad. 
They can work well in domains that are nearly fully observable (such as SMAC \citep{SMAC}) and domains in which remembering history information is not helpful (observations are not very informative or the task is too hard to solve). 
This result has been shown theoretically and empirically \citep{AAAI22,JAIR23}. 

In particular, MADDPG \citep{MADDPG} and COMA \citep{COMA} popularized the idea of using state-based critics using the intuition that this ground truth information that is available during training can help address partial observability. 
In truth, it is typically a bad idea to eliminate partial observability in the critic. A Dec-POMDP typically \emph{is} partially observable and the actor is using partially observable information (i.e., histories). The state-based critic can be biased (i.e., incorrect) because there is a mismatch between the actor and the critic. The actor needs to use history information to choose actions. The state-based critic uses state values instead. As a result, the state values are averaged over the histories that are visited and can't possibly have the correct history values unless there is a one-to-one mapping from states to histories (which would happen in the fully observable case). That is, the critic loses the information about partial observability, which is necessary for the actor to make good choices. 

For example, in the class Dec-Tiger problem \citep{Nair03IJCAI,Book16}, there are only 2 states (the tiger being on the left or on the right) but noisy observations are received when an agent listens (an agent can also choose to open one of the two doors). A state-based critic would only have two values $\jV(\sI 1)$ and $\jV(\sI 2)$ but there are an exponential number of histories that depends on the horizon. As more information is gathered (i.e., the same observation about the tiger's location is heard multiple times) the value of the history, $\jV(\jh)$, should increase (so  $\jV(\jh_{tiger-left-10-times-in-a-row}) > \jV(\jhT 0)$, the initial history). This can't be reflected in the state-based critic. Furthermore, information-gathering actions (i.e., listening in this tiger problem) won't improve the value according to the critic so they shouldn't be taken.  As a result, the state-based critic values will be incorrect in the Dec-Tiger problem and the agents would learn a policy to open one of the doors and hope for a good outcome (assuming the tiger is randomly initialized). 

The updates to Algorithm \ref{alg:IA2CC} for the state-based critic case are straightforward---$\jVmodel(s)$ is used in place of $\jVmodel(\jh)$. Then, the TD error calculation becomes $\delta_t \gets r_t + \gamma \jVmodel(s_{t+1}) - \jVmodel(s_t)$, which is used in the actor and critic updates.

It turns out that history-state critics, those that take both the state and history as input ($\jQ(\jh,s,\ja)$ or $\jV(\jh,s)$ are unbiased and often perform the best \citep{AAAI22,JAIR23,MAPPO}. This result has been theoretically and empirically shown for the general single and multi-agent cases \citep{AAMAS22Baisero,AAAI22,JAIR23} as well as empirically with variants of MAPPO \citep{MAPPO}. 
The intuition for correctness of the history-state critic is that no history information is lost (as it is in the simpler state-based critic). That is, the correct history value function can be (and is) recovered from the history-state value:
$\jQpol(\jh,\ja) = \mathbb{E}_{s\mid \jh} \left[ \jQpol(\jh, s, \ja) \right]$. 
The MAPPO paper also showed that if you can handcraft the features of the history-state critic to retain only the relevant information, the performance can be the highest since the agent doesn't need to learn which information is relevant and which is not. Unfortunately, determining which information is relevant and properly separating it can be difficult to do. 
The algorithmic updates for the history-state critic case are also straightforward---$\jVmodel(\jh,s)$ is used in place of $\jVmodel(\jh)$. The TD error calculation becomes $\delta_t \gets r_t + \gamma \jVmodel(\jh,s_{t+1}) - \jVmodel(\jh,s_t)$, which is used in the actor and critic updates.  

\subsection{Choosing different types of decentralized and centralized critics}
\label{sec:ctde:pg:critics}

While CTDE methods are (by far) the most popular form of MARL, they do not always perform the best. In fact,  decentralized training and execution methods (DTE) \citep{DTE} are actually quite close to CTDE methods. This is because DTE methods typically make the concurrent learning assumption where all the agents learn using the same algorithm, making updates at the same time on the joint data. 
This equivalence is seen explicitly in the policy gradient case as the gradient of the joint update is the same as the decentralized gradient \citep{Peshkin00UAI}. This phenomenon has been also been studied theoretically and empirically with modern actor-critic methods \citep{AAMAS21Luke,JAIR23}. It turns out that while centralized critic actor-critic methods are often assumed to be better than DTE actor-critic methods, they are theoretically the same (with mild assumptions) and often empirically similar (and sometimes worse). Too much centralized information can sometimes be overwhelming, harming scalability of centralized-critic methods and leading to higher variance \citep{MAPPO,JAIR23}. 

The theory assumes learned critics (or sufficient Monte Carlo estimates) but different types of critics may be easier to learn in different cases. 
As a result, it is not straightforward to determine when to use each critic in practice. 
Centralized critics often cause higher variance actor updates since information from the other agents needs to be marginalized out (e.g., through sampling) while information from the other agents is already removed from the decentralized critics. 
Centralized critics can also be more difficult to learn in problems with large numbers of agents or with large action or observation sets, making decentralized critics more scalable. 
Decentralized critics may be difficult to learn due to agents changing their policies (i.e., nonstationarity), making critic estimates stale. 
State-based critics are often the easiest to learn (because no history representation needs to be learned) but are biased in partially observable settings and will have high variance (again, because of the mismatch between the actor and the critic). 
In general, learning a history representation is difficult. Agents need to figure out what history information is relevant and what is not from a noisy and often sparse RL signal. Using a recurrent network (or even a transformer) is usually not sufficient for this task. 
History-state critics further increase the variance in the policy update (because the state information needs to also be marginalized out) but can often be easier to learn than centralized critics with only history information.\footnote{The exact reason why this is the case is unclear and a great topic for future research!} 
As a result the choice of critic is often a bias-variance tradeoff since, in practice, a centralized critic should have a lower bias than decentralized critics with more stable values that are more easily updated when policies change but higher variance because the policy updates need to be averaged over other agents.

There are many different ways of performing CTDE, and current work has only scratched the surface. Studying the differences and similarities between DTE and CTDE variants of algorithms seems like a promising direction for understanding current methods and developing improved approaches for both cases. Determining what centralized information to use and how to best use it is another key question. 
Lastly, to the best of my knowledge, there are no globally optimal model-free MARL methods for Dec-POMDPs. This is surprising since there are optimal \emph{planning} methods where the model is assumed known \citep{Book16} and many globally optimal model-free methods for single-agent RL \citep{SuttonBarto18}. Developing such methods (even for the tabular case) would be interesting and may lead to new, better methods that could be approximated in the deep case.  

\subsection{Methods that combine policy gradient and value factorization}

Several methods combine using a centralized critic with value factorization. 
One notable method is
FACtored Multi-Agent Centralized policy gradients (FACMAC) \citep{facmac} which extends MADDPG to include a QMIX-style factored centralized critic, but since the local Q-values don't need to be used for action selection (the actor does that), the factorization can be nonmonotonic.  Also, rather than sampling the actions of other agents from the off-policy dataset, as MADDPG does, actions for other agents are sampled from the current policies during the actor updates. 
Decomposed Off-Policy policy gradient (DOP) \citep{DOP} uses a factored centralized critic that is a weighted sum of the local Q-values 
and a multi-agent extension of tree backup \citep{TreeBackup} to more efficiently calculate off-policy updates. While a centralized critic is learned, each agent's local Q-value estimate is used to update its actor and with mild assumptions the method is shown to converge to a local optimum even with the simple critic.

\subsection{Other centralized critic methods}

Many other methods use centralized critics. For instance, 
\cite{MAAC} use attention for determining which other agent information to include in the centralized critic. 
Qatten~\citep{Qatten} extends QMIX (and weighted QMIX) to add structure to the mixing network, where the structure is inspired by a linear approximation of the Q-function and one set of weights is leaned using attention. A wide range of other approaches have been developed but are not included in this introduction to the area.

\section{Other forms of CTDE}

While the methods above are the ones that are most widely used, there are several other forms of CTDE. I discuss common approaches and include a note on the many other topics that are not in this text in order to focus on core issues of CTDE. 

\subsection{Adding centralized information to decentralized methods}

As noted in the introduction, decentralized training and execution (DTE) methods are independent learning methods \citep{Claus98AAAI}  where each agent learns on its own using only its own information. More details about decentralized training methods can be found in a companion text \citep{DTE}.  DTE fit well when there is no offline training phase or when scalability is the key factor. DTE methods can be augmented with centralized information to potentially improve performance. These extensions are discussed below. 

\paragraph{Parameter sharing}

Many cooperative MARL methods use parameter sharing. As noted in Section \ref{sec:ctde:pg:mappo}, the idea behind parameter (or weight) sharing is instead of each agent using a different network to estimate the value function or policy, all agents share the same networks. 
The data from each agent can be used to update a single value network for an algorithm such as DRQN, speeding up learning. Agents can still perform differently due to observing different histories and agent indices can be added to increase specialization \citep{Gupta17,Foerster16}. 
While parameter sharing is typically used with homogeneous agents (i.e., those with the same action and observation spaces), it can also be used with heterogeneous agents \citep{terry2020revisiting}.
Since parameter sharing requires agents to share networks during training, it couldn't be used for online training in a decentralized fashion. Nevertheless, decentralized algorithms can be extended to use parameter sharing (such as the case with IPPO). Parameter-sharing implementations may be more scalable than other forms of CTDE, are often simple to implement, and can perform well \citep{Gupta17,COMA,MAPPO}. 

\paragraph{Alternating learning}

A number of methods allow agents to take turns learning. For instance, decentralized learning methods \citep{DTE} can be used but all agents are fixed (not learning) except for one. The learning agent learns until convergence, generating a best-response to the other agent policies. If this process continues until no agents can further improve their polices, the result is a Nash equilibrium \citep{Banerjee12,ma2ql}. 
In limited settings with additional strong assumptions (e.g., deterministic environments, full observability, additional coordination mechanisms to ensure coordinated policies), such methods can potentially converge to an optimal solutions \citep{Lauer00ICML,jiang2023best,jiang2022i2q}. 
While these methods are sometimes called `decentralized' since coordination and communication is need during learning that is not available during execution, I consider them to be CTDE. 

\cite{kuba2022trust} use sequential agent updates (i.e., one agent updates at a time while holding the others fixed) in Heterogeneous-Agent Trust Region Policy Optimisation (HATRPO) and Heterogeneous-Agent Proximal Policy Optimisation (HAPPO). HATRPO and HAPPO build off of TRPO \citep{TRPO} and PPO \citep{PPO} and remove parameter sharing used in MAPPO \citep{MAPPO} (leading to the heterogeneous-agent name).  Using this sequential update scheme, they can theoretically prove monotonic improvement for the fully observable case and the approaches can work well in practice.

\paragraph{Addressing nonstationarity}

Decentralized learning methods face a nonstationary problem due to changing policies of other agents. Some methods try to address this challenge by directly modeling these changes. For example, \cite{Foerster17} propose using importance sampling to correct for the probability differences and decay old data or simpler `fingerprints' (e.g., episode number) to mark the data's age. 
Other methods try to model the other agent learning updates as well \citep{LOLA,COLA}.

\subsection{Decentralizing centralized solutions}

Another potential approach is to learn a centralized solution during training and then (attempt to) decentralize it before execution. This is much harder than it seems. 
A centralized policy can map $\jhS \to \jaS$ without the constraint (in the stochastic case) that $\jpol(\ja| \jh)=\prod_{i \in \agentS} \poli(\ai| \hi)$. 
That is, each agent's policy can depend on \emph{other agent histories}: $\jpol(\ja| \jh)=\prod_{i \in \agentS} \poli(\ai| \jh)$. The centralized policy class is much larger and richer than the decentralized policy class. 

For example, just considering deterministic policies of horizon $\hor$, there would be 
$|\aAS i|^{|\oAS i|^{\hor^{|\agentS|}}}$ possible decentralized policies (assuming all agents have the same size action and observation sets) vs.~$|\jaS|^{\joS^\hor}$ possible centralized policies. 
Plugging in 4 actions per agent, 5 observations, a horizon (history-length) of 10 and 4 agents gives 
${{4^5}^{10}}^4 = 1.024  \times 10^{43}$
vs.~${(4^4)^{(5^4)}}^{10} = 2.56 \times 10^{62520}$ policies. This is a massive difference. 

As a result, most centralized policies will not be directly decentralizeable in the sense that they will be equivalent.\footnote{Note that this is not true in the fully observable case as discussed in Section \ref{sec:fullobs}.}  Policies can only be decentralized when they don't depend on other agent information but centralized agents that make joint action choices based on the information of all agent can often perform much better. In fact, this centralized partially observable case is called a multi-agent POMDP and can be solved using single-agent methods \citep{Book16}. 

As a result of these issues, approaches that decentralize centralized solutions are not common. Nonetheless, there are some instances. For example, 
\cite{liexplicit} attempt to reconstruct the global information using a reconstruction loss, other methods attempt to mimic a centralized controller \citep{lin2022decentralized}, and others allow communication during training but reduce or remove it during execution \citep{Foerster16,Wang2020}.
FOP (for Factorizes the Optimal joint Policy) \citep{FOP} makes the assumption that centralized solutions are decentralizeable and then seeks to learn solutions using a max entropy formulation along with a value factorization scheme that is similar to  methods in Section \ref{sec:ctde:vff}. While FOP (and other methods) could be applied in general Dec-POMDP settings, they are likely to perform poorly when the centralized solution is sufficiently different from the decentralized solution.

\subsection{Topics not discussed}

There are many other issues that are important to CTDE but are not discussed in this text, including exploration and communication as well as other topics such as hierarchical methods, role decomposition, ad-hoc (or zero-shot) coordination, and multi-task approaches. While these (and other) topics are very important, I do not include them for the sake of brevity and simplicity.

\section{Acknowledgements}

I thank Andrea Baisero, Shuo Liu, and the other members of my Lab for Learning and Planning in Robotics (LLRP) for reading my (very) rough drafts and providing comments that helped improve the document. 

\bibliographystyle{abbrvnat}
\bibliography{theBib}

\end{document}